\documentclass[10pt,twocolumn,letterpaper]{article}

\usepackage{cvpr}              

\usepackage[utf8]{inputenc}
\usepackage{amssymb}      
\usepackage{pifont}  

\newcommand{\xmark}{\ding{55}}

\usepackage{array} 
\newcolumntype{L}[1]{>{\raggedright\arraybackslash}m{#1}} 
\newcolumntype{C}[1]{>{\centering\arraybackslash}m{#1}}   

\usepackage{booktabs,multirow}

\usepackage[most]{tcolorbox}

\newtcolorbox{promptbox}[1][]{
  enhanced, breakable,
  colback=gray!4,        
  colframe=black!30,     
  boxrule=0.5pt, arc=2pt,
  left=8pt, right=8pt, top=6pt, bottom=6pt,
  fonttitle=\bfseries,   
  title={#1}
}

\definecolor{cvprblue}{rgb}{0.21,0.49,0.74}
\usepackage[pagebackref,breaklinks,colorlinks,allcolors=cvprblue]{hyperref}

\def\confName{CVPR}
\def\confYear{2026}

\title{Beyond Global Similarity: Towards Fine-Grained, Multi-Condition Multimodal Retrieval}

\author{
Xuan Lu$^{1,2,3}$\thanks{Equal contribution} \quad
Kangle Li$^{1,2}$\footnotemark[1] \quad
Haohang Huang$^{3}$ \quad
Rui Meng\thanks{Working at Google Cloud AI Research.} \quad
Wenjun Zeng$^{2,3}$ \quad
Xiaoyu Shen$^{2,3}$\thanks{Corresponding author.} \\
\vspace{1mm} \\
$^{1}$Shanghai Jiao Tong University \\
$^{2}$Institute of Digital Twin, Eastern Institute of Technology, Ningbo \\
$^{3}$Ningbo Key Laboratory of Spatial Intelligence and Digital Derivative \\
{\tt\small lux1997@sjtu.edu.cn} \quad {\tt\small xyshen@eitech.edu.cn}
}

\begin{document}
\maketitle
\begin{abstract}
Recent advances in multimodal large language models (MLLMs) have substantially expanded the capabilities of multimodal retrieval, enabling systems to align and retrieve information across visual and textual modalities. Yet, existing benchmarks largely focus on coarse-grained or single-condition alignment, overlooking real-world scenarios where user queries specify multiple interdependent constraints across modalities.
To bridge this gap, we introduce \textbf{MCMR} (Multi-Conditional Multimodal Retrieval): a large-scale benchmark designed to evaluate fine-grained, multi-condition cross-modal retrieval under natural-language queries. MCMR spans five product domains: upper and bottom clothing, jewelry, shoes, and furniture. It also preserves rich long-form metadata essential for compositional matching. Each query integrates complementary visual and textual attributes, requiring models to jointly satisfy all specified conditions for relevance.
We benchmark a diverse suite of MLLM-based multimodal retrievers and vision–language rerankers to assess their condition-aware reasoning abilities. Experimental results reveal: (i) distinct modality asymmetries across models; (ii) visual cues dominate early-rank precision, while textual metadata stabilizes long-tail ordering; and (iii) MLLM-based pointwise rerankers markedly improve fine-grained matching by explicitly verifying query–candidate consistency.
Overall, MCMR establishes a challenging and diagnostic benchmark for advancing multimodal retrieval toward compositional, constraint-aware, and interpretable understanding. Our code and dataset is available at \url{https://github.com/EIT-NLP/MCMR}
\end{abstract}
\section{Introduction}

Multimodal retrieval systems are designed to align and retrieve information across heterogeneous and composite modalities such as text, images, or their interleaved combinations~\cite{shen2022lowresourcedenseretrievalopendomain,shen2023neural,liu2023bidirectionaltrainingcomposedimage,wang2024crossmodalretrievalsystematicreview,abootorabi2025ask}. Earlier models such as CLIP~\cite{2021clip}, ALIGN~\cite{jia2021scaling}, and BLIP~\cite{li2022blip} are trained on holistic image–caption pairs through contrastive learning and have been widely deployed in web search~\cite{mm-embed,jia2025visualwebinstruct,lu2025rethinking}, e-commerce~\cite{shen2023xpqa,zhang2025research}, and recommendation systems~\cite{multimodalrecommender,han2025lemurlargescaleendtoend}.  However, captions usually provide only holistic and generic descriptions of visual content, so these models tend to emphasize global semantic consistency over fine-grained cross-modal understanding.
\begin{figure}
    \centering
    \includegraphics[width=0.99\linewidth]{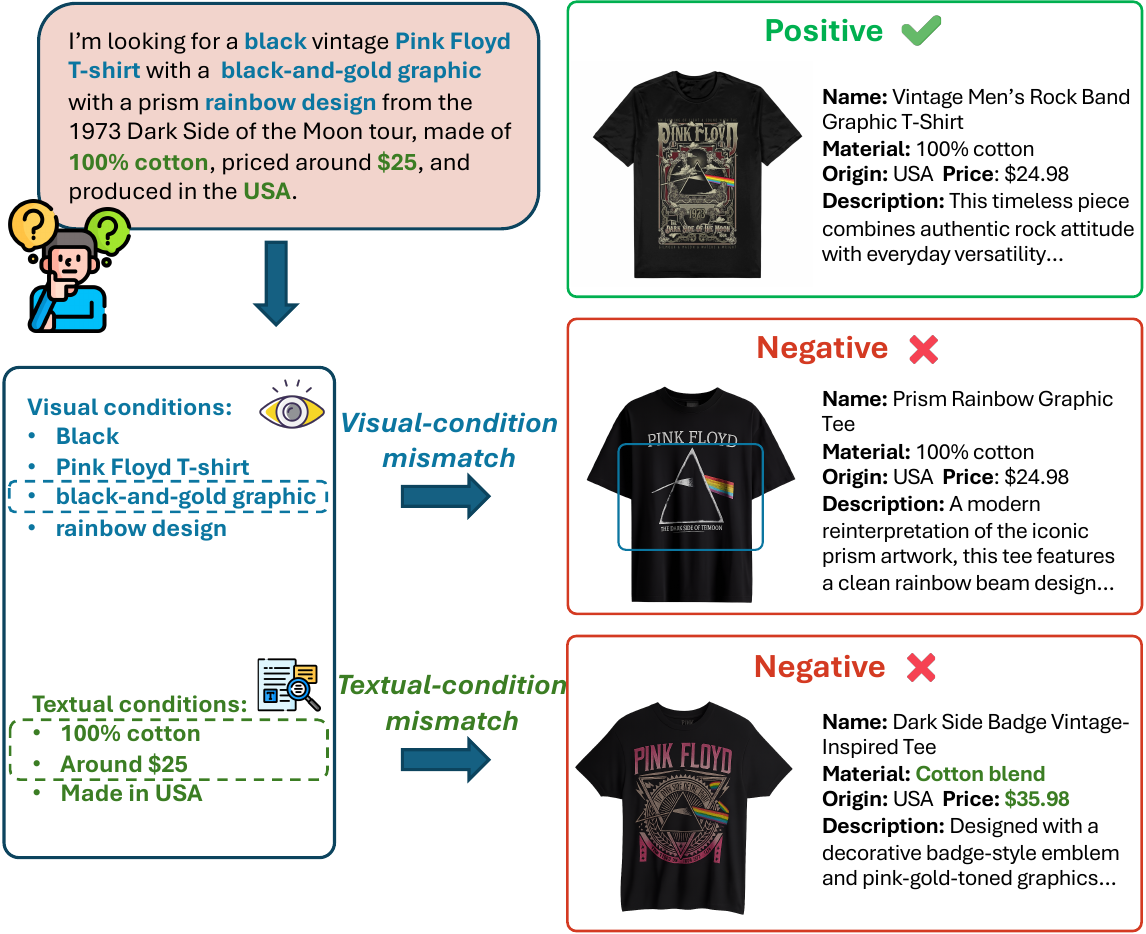}
    \caption{\small Multi-condition multimodal retrieval driven by natural-language queries with fine-grained visual and textual constraints.}
    \label{fig:introduction}
    \vspace{-4pt}
\end{figure}
In recent years, the rapid advancement of multimodal large language models (MLLMs) has begun to reshape this landscape by enabling retrieval under open-ended natural language instructions~\cite{zhou2024beyond,zhang2025gme,lu2026tools,lu2025rethinking,jiangvlm2vec}. Building on this capability, representative methods such as VLM2Vec~\cite{jiangvlm2vec}, MM-Embed~\cite{mm-embed}, and GME~\cite{zhang2025gme} further broaden the scope of multimodal retrieval by producing unified embeddings that capture instruction-conditioned semantics. This evolution marks a shift from static global alignment toward more expressive and flexible retrieval paradigms.

Despite this progress, existing benchmarks rarely satisfy \emph{three} crucial properties required for complex multimodal retrieval simultaneously: (i) \textbf{fine-grained} attribute reasoning, (ii) \textbf{multi-condition} queries, and (iii) \textbf{cross-modal evidence}, where different conditions must be grounded in both images and text. Classical image--text datasets such as MS-COCO~\cite{lin2014microsoft} and Flickr30K~\cite{DBLP:conf/iccv/PlummerWCCHL15} primarily evaluate coarse, global alignment between an image and a caption: each caption is treated as a single holistic description, and relevance does not depend on satisfying multiple independent constraints. 
Fine-grained benchmarks in fashion and grounding, such as FashionIQ~\cite{DBLP:conf/iccv/HanWHZZLZD17}, and CIRR~\cite{Liu2021ImageRO}, introduce localized edits or relative adjustments (e.g., \textit{“shorter sleeves,” “same style but brighter color’’}) and rely on reference images to anchor these modifications. While these datasets provide finer detail than global captioning benchmarks, they still center around \emph{single} visual edits and remain effectively single-modality, as most attributes can be verified from images alone. In parallel, multi-condition retrieval has been explored in text-only settings~\cite{lumulticonir,zhangssrb}, where queries enumerate several constraints but all evidence resides in text, eliminating the need to integrate heterogeneous visual and textual cues. Existing benchmarks capture either fine-grained detail or multi-condition structure, but not both across modalities.
Recent efforts such as MERIT~\cite{chow2025merit} expand the retrieval setting through multilingual, interleaved text–image queries that combine reference images with short textual fragments. In MERIT, many query conditions are formulated relative to exemplar images (e.g., \textit{“same style as Product 1 and same color as Product 2”}), which frames the task around visual comparisons to provided examples rather than independently specified attribute constraints. The formulation also presumes that users can supply representative reference images at query time, whereas in many real-world search scenarios, users more commonly articulate their needs directly through natural-language descriptions. In addition, MERIT does not explicitly separate attributes that require visual grounding from those present only in textual metadata, making it challenging to analyze how models balance evidence across modalities.

To address these limitations, we introduce \textbf{MCMR} (Multi-Conditional Multimodal Retrieval), a large-scale benchmark explicitly designed around the three axes above: \emph{fine-grained}, \emph{multi-condition}, and \emph{cross-modal}. As shown in Fig.~\ref{fig:introduction}, each query is a natural-language description that combines multiple compositional constraints over visual and textual attributes, and a candidate is relevant only if \emph{all} conditions are satisfied. Crucially, MCMR enforces a dual-evidence design in which some attributes are only inferable from the image (e.g., \textit{specific graphic layout or texture}) while others are only supported by long-form textual metadata (e.g., \textit{material, fit, or manufacturing details}). This makes it impossible to solve the task from a single modality and directly tests a model's ability to integrate complementary evidence across modalities.

We benchmark a suite of representative multimodal retrievers under a unified protocol to assess their capacity for fine-grained, condition-aware reasoning, and further introduce MLLM-based pointwise rerankers to examine whether large vision--language models can bridge fine-grained semantic gaps through one-to-one, constraint-aware relevance estimation. Our experiments reveal a persistent challenge in multi-condition multimodal retrieval. Modality ablation analysis uncovers clear asymmetries in modality dependence: models such as GME~\cite{zhang2025gme} and LamRA~\cite{liu-etal-2025-lamra} remain relatively robust when textual metadata is removed, suggesting a stronger reliance on visual features, whereas models like LLaVE~\cite{lan-etal-2025-llave} suffer substantial degradation under image-only settings. Further analysis shows that visual cues dominate top-rank accuracy, textual metadata stabilizes long-tail ordering, and MLLM-based rerankers significantly improve fine-grained matching by explicitly verifying each query--candidate pair. Together, these findings expose the limitations of current systems on truly multi-condition, cross-modal retrieval and highlight the need for architectures that integrate compositional reasoning without sacrificing scalability.

Our main contributions are summarized as follows:

\begin{itemize}
\item \textbf{Benchmark.} We present \textbf{MCMR}, a large-scale benchmark that jointly satisfies three desiderata---fine-grained attributes, multi-condition queries, and cross-modal evidence---for evaluating multimodal retrieval under natural-language queries.

\item \textbf{Comprehensive Evaluation.} We conduct a comprehensive study across representative multimodal retrievers and MLLM-based pointwise rerankers, revealing systematic modality asymmetries and fine-grained reasoning gaps under controlled compositional settings.

\item \textbf{Findings.} Our analysis shows that current retrievers struggle to satisfy multiple heterogeneous cross-modal constraints simultaneously, while MLLM-based reranking markedly improves precision---highlighting the need for scalable, constraint-aware retrieval architectures.
\end{itemize}

\section{Related Works}

\paragraph{Multimodal Retrieval Models.}

Multimodal retrieval aims to align heterogeneous modalities—most commonly images and text—within a unified semantic space, enabling cross-modal search and reasoning.
A dominant paradigm encodes each modality using pre-trained vision–language models such as CLIP~\cite{2021clip}, BLIP~\cite{li2022blip}, and ALIGN~\cite{jia2021scaling}, and measures similarity through cosine distance in the shared embedding space. Building on these dual-encoder backbones, task-agnostic retrievers typically adopt separate encoding for images and text followed by late fusion to reconcile the two modalities. Approaches such as UniVL-DR~\cite{liu2023universal} and UniIR~\cite{wei2023uniir} normalize heterogeneous datasets and map them into a shared space with CLIP- or BLIP-like encoders, then apply calibrated fusion at inference to balance modality contributions across retrieval tasks.
Another line of research attaches lightweight visual adapters to strong text embedding models, allowing image–text pairs to be encoded compositionally without retraining the full model. Representative examples include VISTA~\cite{zhou-etal-2024-vista} and MARVEL~\cite{zhou-etal-2024-marvel}, which inject visual information into text encoders for unified multimodal representations.
More recently, multimodal large language models (MLLMs) have been adapted into general-purpose retrievers by directly pooling hidden states from the final transformer layers as embeddings, as shown in E5-V~\cite{jiang2024e5v} and VLM2Vec~\cite{jiangvlm2vec}. This MLLM-as-embedding paradigm enhances transferability, unifies text–text and cross-modal retrieval, and better aligns with retrieval-augmented generation pipelines.
Across both families—dual encoders and MLLM-based retrievers—models are typically optimized for global semantic alignment and often exhibit distinct modality dependencies: some rely more on visual appearance, while others depend heavily on textual semantics. This asymmetry is one of the core aspects we investigate empirically in our study.

\paragraph{Multimodal Retrieval Benchmarks.}
Early benchmarks for multimodal retrieval emphasize coarse-grained or single-condition alignment between images and captions.
Datasets such as MS-COCO~\cite{lin2014microsoft} and Flickr30K~\cite{DBLP:conf/iccv/PlummerWCCHL15} define one-to-one correspondences between captions and images, focusing on global similarity rather than compositional reasoning.
Domain-specific datasets, including VisualNews~\cite{DBLP:conf/emnlp/LiuWWO21} for news image–headline alignment and Fashion200k~\cite{DBLP:conf/iccv/HanWHZZLZD17} for product retrieval, extend this paradigm to specialized settings but still operate at a holistic level.
Beyond natural images, document-centric suites such as the DocVQA family processed by ViDoRe~\cite{faysse2024colpali} cast retrieval as selecting the correct text-rich document image given a natural-language question.
Subsequent “compositional” retrieval benchmarks shift the focus from global similarity to targeted attribute edits~\cite{zhou2024beyond}.
FashionIQ~\cite{DBLP:conf/cvpr/WuGGARGF21} and CIRR~\cite{Liu2021ImageRO} introduce the idea of modifying a reference image according to a short textual instruction—such as retrieving \textit{“the same shirt but in red”}—to test whether models can incorporate localized attribute changes.
In parallel, fused-modal benchmarks such as WebQA~\cite{DBLP:conf/cvpr/ChangCNGSB22} and EDIS~\cite{DBLP:conf/emnlp/LiuFFCW23} integrate textual and visual evidence in open-domain retrieval, while OVEN~\cite{DBLP:conf/iccv/HuLCKJLTC23} and INFOSEEK~\cite{DBLP:conf/emnlp/ChenHLSCRC23} pair image-conditioned questions with Wikipedia passages or titles to evaluate retrieval under mixed-modality supervision.
Further extensions such as ReMuQ~\cite{luo-etal-2023-end}, OKVQA~\cite{DBLP:conf/cvpr/MarinoRFM19}, and LLaVA-style conversational datasets~\cite{lin-etal-2024-preflmr} broaden the query and candidate modalities to support question answering and dialogue-based retrieval.
More recently, MERIT~\cite{chow2025merit} moves beyond single-reference queries by introducing multilingual and interleaved multimodal inputs that can reference multiple images and text fragments simultaneously, enabling more complex condition specification.
\begin{table}[!htbp]
\centering
\scalebox{1.05}{ 
\begin{tabular}{lccccc}
\toprule
Benchmark & Q & T & DE & MA & LM \\
\midrule
MS-COCO~\cite{lin2014microsoft}     & T   & I   & \xmark & \xmark & \xmark \\
Flickr30K~\cite{DBLP:conf/iccv/PlummerWCCHL15}   & T   & I   & \xmark & \xmark & \xmark \\
FashionIQ~\cite{DBLP:conf/cvpr/WuGGARGF21}   & T+I & I   & \xmark & \xmark & \xmark \\
CIRR~\cite{Liu2021ImageRO}        & T+I & I   & \xmark & \xmark & \xmark \\
WebQA~\cite{DBLP:conf/cvpr/ChangCNGSB22}       & T   & T+I & \xmark & \xmark & \xmark \\
EDIS~\cite{DBLP:conf/emnlp/LiuFFCW23}        & T   & T+I & \xmark & \xmark & \xmark \\
MERIT~\cite{chow2025merit}       & T+I & T+I & \checkmark & \checkmark & \xmark \\
MultiConIR~\cite{lumulticonir}  & T   & T   & \xmark & \checkmark & \checkmark \\
\textbf{MCMR (ours)} & T & T+I & \checkmark & \checkmark & \checkmark \\
\bottomrule
\end{tabular}
}
\caption{
Comparison of representative multimodal retrieval benchmarks. 
Q = query modality, T = target modality, DE = Dual-Evidence, 
MA = Multi-Attribute , LM = Long-form Metadata. 
Entries in Q/T use T for text, I for image, and T+I when both modalities are available. 
MCMR uniquely supports dual-evidence grounding, multi-attribute conditions, and long-form metadata.
}
\label{tab:benchmark_comparison}
\end{table}
\section{Multimodal Multiconditional Retrieval}
\label{sec:method}
We introduce MCMR, a multimodal product retrieval dataset comprising 10,400 products across five retrieval scenarios. The atomic unit is a product instance with an image and a long-form textual description, where the two modalities contribute complementary information. Tab.~\ref{tab:benchmark_comparison} summarizes how MCMR compares with prior multimodal retrieval benchmarks in terms of query–target modality, evidence structure, and metadata richness.

\begin{figure*}
    \centering
    \includegraphics[width=0.95\linewidth]{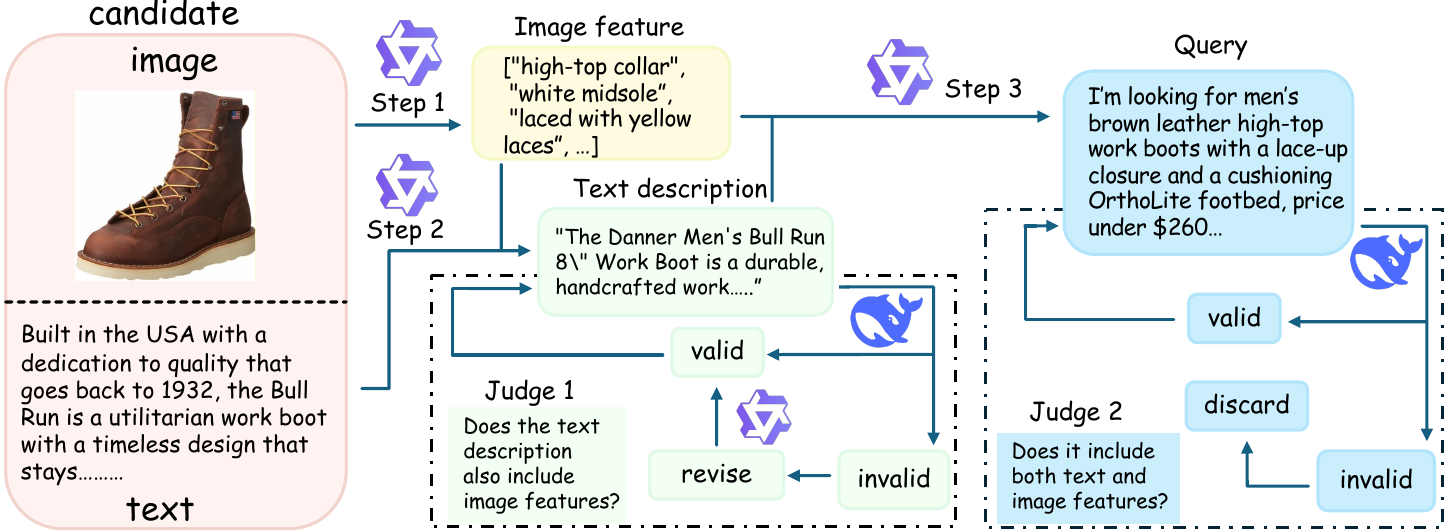}
    \caption{Construction pipeline of MCMR. Visual and textual features are extracted by Qwen-VL and Qwen-Instruct, validated through two judging stages, and composed into queries for fine-grained retrieval evaluation.}
    \label{fig:placeholder}
\end{figure*}
\begin{table*}[ht]
\centering
\setlength{\tabcolsep}{5pt} 
\begin{tabular}{lrrrrrr rrr}
\toprule
& \multicolumn{6}{c}{Domain Distribution (Counts)} & \multicolumn{3}{c}{Token Statistics (Tokens)} \\
\cmidrule(r){2-7} \cmidrule(l){8-10}
& Upper & Bottom & Shoe & Jewelry & Furniture & \textbf{Total}  & max &min & Avg. \\
\midrule
Queries    & 991 & 803 & 847 & 602 & 754 & \textbf{3\,997}  & 57.00  & 25.00 & 35.86 \\
Candidates & 29\,986 & 29\,514 & 24\,997 & 5\,491 & 14\,993 & \textbf{104\,981} & 269.00 & 54.00 & 190.94 \\
\bottomrule
\end{tabular}
\caption{Domain-wise counts and token-length statistics (max/min/avg) for queries and candidates.}
\label{tab:combined_fullwidth_statistics}
\end{table*}

\subsection{Data Collection and Preprocessing}
Built on the Amazon Reviews (2023) corpus~\cite{hou2024bridging}, MCMR spans five product domains—upper and bottom clothing, jewelry, shoes, and furniture—while preserving long-form product metadata essential for fine-grained multimodal retrieval. To ensure quality and reproducibility, we construct the dataset via a multi-stage pipeline that combines automated filtering with manual verification, progressively refining from broad candidate harvesting to fine-grained quality control.
We follow three principles: broad coverage, high quality, and cross-modal consistency.
Starting from Amazon Reviews metadata, we retain items with both resolvable images and long-form textual descriptions in the four target domains. Data curation proceeds in three stages:

\textbf{(1) Attribute normalization.}
We standardize structured attributes by unifying units and currencies, normalizing dates, and aligning controlled vocabularies for materials and sizes to create a unified representation across domains.

\textbf{(2) Quality filtering and de-duplication.}
We apply  filtering and near-duplicate detection based on text length, image resolution, and aspect ratio, and perceptual or embedding similarity. Records that are low-quality, incomplete, or redundant are removed. Direct identifiers such as ASINs and URLs are masked or excluded to prevent leakage.

\textbf{(3) Complementarity constraint.}
Each item must include at least one \emph{text-only} and one \emph{image-only} fine-grained attribute, ensuring that both modalities contribute unique evidence rather than allowing the task to be solved from a single modality.

\subsection{Dataset Construction Pipeline}

To balance cost and fidelity, we adopt a cooperative pipeline pairing efficient mid-sized models for large-scale generation with stronger models for validation and refinement. The goal is to generate \emph{multi-condition, cross-modal, and verifiable} natural-language queries grounded in complementary visual and textual evidence.\footnote{The prompts used for constructing MCMR are provided in the supplementary material.}The overall construction pipeline is illustrated in Fig.~\ref{fig:placeholder}.

\paragraph{Image-side structured expansion.}
From each product image, Qwen2.5-VL-32B-Instruct produces a structured, evidence-grounded summary with a category label and \emph{image-only} attributes (e.g.,color, texture, structural details, shape, structure). The generation strictly excludes functional or speculative content, forming the visual evidence layer used for (i) complementarity verification and (ii) query construction.

\paragraph{Text-side structured expansion.}
Product titles, descriptions, and feature lists are converted into structured text profiles using a JSON extraction template. Required fields include attributes\_text\_only and category\_text; optional fields (e.g., price, release year, material, style, brand) are added when present. Identifiers are sanitized, and brand mentions are allowed only if they co-occur with image-only attributes. Schema-aware merging preserves canonical metadata while appending structured fields for multimodal alignment.

\paragraph{Textual description generation.}
We employ Qwen3-32B-Instruct to generate concise (80–120 words), catalog-style textual summaries based solely on text metadata, explicitly excluding any visual descriptors. A validator–editor loop using DeepSeek-R1-Distill-Qwen-32B detects cross-modal leakage and enforces consistency, with human spot checks ensuring robustness.

\paragraph{Query generation.}
Conditioned on both the image attributes and textual summaries, Qwen3-32B-Instruct generates first-person, multi-condition queries combining complementary constraints from both modalities. Numeric and temporal information is normalized, identifiers are masked, and a neutral “shopper” tone is maintained. Domain-specific prompt variants (for clothing, jewelry, shoes, and furniture) guide attribute selection—e.g.,\emph{fabric/fit/care} for clothing or \emph{gemstone/cut/setting} for jewelry—to preserve semantic diversity.

\paragraph{Query verification.}
DeepSeek-R1-Distill-Qwen-32B serves as an independent verifier assessing cross-modal coverage and numeric/time consistency. Failing or borderline cases are regenerated, and a small subset of accepted samples is manually checked. 
To further validate quality, we conduct a 100-sample human study: one annotator writes natural-language queries from each product’s image and metadata, and a second annotator evaluates, under double-blind conditions, both human-written and generated queries on attribute correctness, cross-modal coverage, and naturalness (5-point Likert scale). The two types of queries achieve comparable average scores (4.33 vs.\ 4.41) and similar preference rates (47\% vs.\ 49\%), confirming that generated queries closely match human-authored ones with minimal verification overhead\footnote{Details of human validation are provided in supplementary material.}.
Tab.~\ref{tab:combined_fullwidth_statistics} presents key statistics of the MCMR dataset, including the number of queries and candidates as well as their text token lengths.

\section{Experiments}
\label{sec:experiments}
This section first presents the experimental settings in § 4.1, followed by the main results in § 4.2, which include overall comparisons of retrieval models. § 4.3 then presents the reranking performance of MLLMs. All experiments are conducted on two NVIDIA A100 GPUs (80 GB). 

\subsection{Baselines and Metrics}

We evaluate five representative multimodal retrievers and five MLLM-as-Rerankers under identical preprocessing and inference settings:
\vspace{2pt}
\begin{itemize}
\item \textbf{Multimodal retrievers:} GME-Qwen2-VL-7B~\cite{zhang2025gme}, LLaVE-7B~\cite{lan-etal-2025-llave}, VLM2Vec~\cite{jiangvlm2vec}, LamRA-Ret-Qwen2.5-VL-7B~\cite{liu-etal-2025-lamra}, and CORAL~\cite{chow-etal-2025-merit-coral}.
\item \textbf{MLLM-as-Rerankers:} Qwen2.5-VL-32B, Qwen2.5-VL-7B, Qwen3-VL-4B, Qwen3-VL-8B~\cite{Qwen2.5-VL}, InternVL3-8B-Instruct~\cite{chen2024internvl}, Qwen3-VL-Reranker-8B~\cite{qwen3vlembedding}, and lychee-reranker-mm~\cite{lychee}.

\end{itemize}
\vspace{2pt}

For retrieval, we embed candidate items using the model’s fused image–text interface,
while queries are encoded from text only. All systems operate in zero-shot mode using publicly
released checkpoints. For reranking, we take the top-50 candidates returned by the strongest first-stage retriever and apply vision–language MLLMs as pointwise rerankers. Each query–candidate pair is evaluated independently: the reranker receives the textual query together with the candidate’s image and metadata, and produces a binary relevance judgment (true/false) through a unified prompt. The model’s normalized logit score for the “true” token is used as the relevance score, and candidates are sorted by this score to obtain the final ordering. This pairwise evaluation complements the first-stage retriever by providing fine-grained, cross-modal relevance estimation.

We report three widely used IR metrics: Recall@K, nDCG@K, and MRR@10.
Recall@K measures the proportion of relevant items retrieved within the top-$K$ ($K\in\{1,5,10,50,100\}$), reflecting the system’s ability to surface correct results under multi-condition constraints.
nDCG@K evaluates ranking quality by weighting relevant documents according to their positions, emphasizing fine-grained ordering within the top results.
MRR@10 summarizes early-ranking performance through the average inverse rank of the first relevant item within the top-10 positions.

\subsection{Main Results}

We evaluate multimodal retrievers on MCMR under three candidate-visibility regimes: fused (image+text), image-only, and text-only. 
Complete results appear in Tab.~\ref{tab:MCMR-bytype-subtables}.

\paragraph{Fused Candidates (Image + Text).}
As shown in Tab.~\ref{tab:MCMR-bytype-subtables}, when both visual and textual metadata are available, overall accuracy remains moderate: mainstream retrievers achieve only 18–27\% Recall@1, while \emph{VLM2Vec} performs notably worse at 1.83\%. Despite this, most models eventually retrieve the correct item at larger cutoffs. The best Recall@10 reaches 53.34\% (CORAL), and the strongest long-range performance—78.64\% Recall@100—is achieved by LLaVE, indicating that relevant items are often found but not ranked near the top.
This widening gap between early-rank and long-cutoff retrieval reveals a consistent pattern across systems: models are reasonably capable at coarse retrieval but struggle with fine-grained ordering under multi-condition constraints. This discrepancy underscores substantial headroom for downstream rerankers that can refine ordering once the correct candidate is retrieved.

\paragraph{Effect of Text Removal (Image-Only Candidates).}
Removing textual metadata while keeping images strongly impacts early-rank accuracy. For several models, R@1/5/10 drops sharply, and in some cases top-1 accuracy falls below 1\%. In contrast, a few visually strong encoders (e.g., GME) maintain competitive scores and sometimes even gain at R@1, indicating that some systems can satisfy multiple conditions using visual cues alone. Performance becomes more stable at larger cutoffs: R@50/100 remains relatively high, and the degradation in nDCG@100 is far smaller than the drop at nDCG@10. This pattern suggests that text removal primarily reduces fine-grained discrimination but does not severely hinder the ability to retrieve the correct item somewhere within the candidate set.

\paragraph{Effect of Image Removal (Text-Only Candidates).}
Removing images and relying solely on textual metadata leads to a clear drop in accuracy across all metrics compared with fused candidates. Early-rank retrieval is particularly weak: the best Recall@1 reaches only 12.98\% (MM-EMBED), with most models in the 7–12\% range. Mid- and long-cutoff performance also decreases—Recall@50/100 remain well below fused results, topping out around 53–62\%.
Text-only retrieval is also consistently weaker than image-only retrieval. Visually strong models such as GME-Qwen2VL show substantial decreases from 51.10\% to 29.60\% at Recall@10 and from 78.86\% to 57.50\% at Recall@100. These results highlight the predominantly visual nature of MCMR’s discriminative cues. Text functions mainly as a complementary signal that enhances alignment when combined with images, while the strongest performance consistently arises when both modalities are available.

\begin{table*}[t]
\centering
\scriptsize
\setlength{\tabcolsep}{3pt}

\begin{subtable}{\linewidth}
\centering
\subcaption{Fused (image+text)}
\resizebox{\linewidth}{!}{
\begin{tabular}{l c r r r r r r r r r r r r}
\toprule
model & size & R@1 & R@5 & R@10 & R@50 & R@100 & MRR & N@1 & N@5 & N@10 & N@50 & N@100 \\
\midrule
LLaVE            & 7B & \underline{24.99} & \underline{43.85} & \underline{53.13} & \textbf{72.01} & \textbf{78.64} & \underline{33.15} & \underline{24.99} & \underline{34.88} & \underline{37.88} & \underline{42.11} & \underline{43.19} \\
GME-Qwen2VL      & 7B & 21.23 & 38.20 & 45.74 & 64.71 & 73.52 & 28.35 & 21.23 & 30.06 & 32.48 & 36.66 & 38.08 \\
LamRA-Qwen2.5VL  & 7B & 17.96 & 34.99 & 43.30 & 64.36 & 73.24 & 25.27 & 17.96 & 26.85 & 29.53 & 34.25 & 35.69 \\
MM-EMBED         & 8B & 21.74 & 39.58 & 47.91 & 66.22 & 74.16 & 29.35 & 21.74 & 31.05 & 33.75 & 37.82 & 39.11 \\
CORAL            & 3B & \textbf{26.57} & \textbf{46.69} & \textbf{53.34} & \underline{70.90} & \underline{77.73} & \textbf{34.94} & \textbf{26.57} & \textbf{37.20} & \textbf{39.35} & \textbf{43.27} & \textbf{44.37} \\
VLM2Vec          & 4B &  1.83 &  4.88 &  7.03 & 14.38 & 18.96 &  3.11 &  1.83 &  3.33 &  4.02 &  5.63 &  6.38 \\
\bottomrule
\end{tabular}}
\end{subtable}

\vspace{0.8em}

\begin{subtable}{\linewidth}
\centering
\subcaption{Image-only}
\resizebox{\linewidth}{!}{
\begin{tabular}{l c r r r r r r r r r r r r}
\toprule
model & size & R@1 & R@5 & R@10 & R@50 & R@100 & MRR & N@1 & N@5 & N@10 & N@50 & N@100 \\
\midrule
LLaVE            & 7B &  0.90 &  2.53 &  3.93 &  9.48 & 13.68 &  1.67 &  0.90 &  1.52 &  2.19 &  3.39 &  4.06 \\
GME-Qwen2VL      & 7B & \textbf{21.79} & \textbf{41.30} & \textbf{51.10} & \textbf{71.36} & \textbf{78.86} & \textbf{30.13} & \textbf{21.79} & \textbf{31.91} & \textbf{35.08} & \textbf{39.63} & \textbf{40.86} \\
LamRA-Qwen2.5VL  & 7B & \underline{18.05} & \underline{36.25} & \underline{43.30} & \underline{66.83} & \underline{76.73} & \underline{25.96} & \underline{18.05} & \underline{27.63} & \underline{30.51} & \underline{35.35} & \underline{36.96} \\
MM-EMBED         & 8B & 13.23 & 28.15 & 35.68 & 57.29 & 67.00 & 19.72 & 13.23 & 21.06 & 23.50 & 28.30 & 29.89 \\
CORAL            & 3B & 11.51 & 25.99 & 33.53 & 54.72 & 64.15 & 17.83 & 11.51 & 19.11 & 21.54 & 26.19 & 27.72 \\
\bottomrule
\end{tabular}}
\end{subtable}

\vspace{0.8em}

\begin{subtable}{\linewidth}
\centering
\subcaption{Text-only}
\resizebox{\linewidth}{!}{
\begin{tabular}{l c r r r r r r r r r r r r}
\toprule
model & size & R@1 & R@5 & R@10 & R@50 & R@100 & MRR & N@1 & N@5 & N@10 & N@50 & N@100 \\
\midrule
LLaVE            & 7B & \underline{11.95} & \underline{23.38} & 29.43 & 48.02 & 56.23 & \underline{16.83} & \underline{11.95} & \underline{17.85} & \underline{19.80} & \underline{23.92} & \underline{25.25} \\
GME-Qwen2VL      & 7B & 10.62 & 22.65 & \underline{29.60} & \underline{48.55} & \underline{57.50} & 16.00 & 10.62 & 16.95 & 19.21 & 23.42 & 24.87 \\
LamRA-Qwen2.5VL  & 7B &  7.28 & 16.21 & 22.31 & 38.53 & 48.49 & 11.27 &  7.28 & 16.21 & 13.85 & 17.36 & 18.98 \\
MM-EMBED         & 8B & \textbf{12.98} & \textbf{26.88} & \textbf{34.50} & \textbf{53.66} & \textbf{62.37} & \textbf{18.94} & \textbf{12.98} & \textbf{20.15} & \textbf{22.61} & \textbf{26.86} & \textbf{28.67} \\
CORAL            & 3B &  8.58 & 17.10 & 22.88 & 39.30 & 47.73 & 12.37 &  8.58 & 12.98 & 14.83 & 18.43 & 19.80 \\
\bottomrule
\end{tabular}}
\end{subtable}
\caption{Retrieval on MCMR under three candidate-visibility regimes (fused/image-only/text-only). Queries are text-encoded; we report R@K (K=1,5,10,50,100), NDCG@K, and MRR@10.Values are percentages.}
\label{tab:MCMR-bytype-subtables}
\end{table*}

\paragraph{Gains from Pointwise Reranking}

We apply a second-stage pointwise reranker to the fixed top–50 candidates returned by the LLaVE-7B, which achieves the highest Recall@50 under the fused setting (72.01). Each reranker is a vision–language MLLM that independently assesses the relevance of a text query paired with a fused image–text candidate. The model outputs a normalized probability for the token \texttt{True} at its first answer position, which we use as the relevance score $p(\mathrm{True})$; candidates are then sorted by this score, with ties broken by their original retrieval rank.
Results shown in Fig. \ref{tab:reranking}, across the fixed top–50 pool, pointwise rerankers provide substantial improvements in early-rank ordering, with nDCG@1 scores consistently in the 70–80 range. lychee-reranker-mm achieves the strongest performance at every cutoff—92.35 / 93.41 / 94.42 / 94.86 for nDCG@1/5/10/50—followed by internVL-8B and Qwen3-VL-Reranker-8B, while the Qwen3-VL-based models lag slightly behind. Parameter count alone does not predict ranking ability; architectural design and vision–text grounding appear more consequential for pairwise relevance estimation.
Performance gaps are largest at nDCG@1 and narrow toward nDCG@50, indicating that MLLM rerankers concentrate their gains at the very top while long-range ordering remains relatively stable.  The strong one-to-one relevance judgments of MLLM rerankers, contrasted with the modest Recall@1 achieved by first-stage retrievers, reveal a clear gap: current retrieval models struggle with fine-grained, multi-condition semantic reasoning, whereas generative vision–language models can bridge this gap when evaluating each pair independently. This contrast also validates the design of MCMR as a benchmark: \emph{it effectively exposes weaknesses in cross-modal fusion and fine-grained retrieval, providing a meaningful testbed for evaluating models’ multimodal alignment and condition-aware reasoning abilities.}



\begin{table}[t]
\centering
\resizebox{\columnwidth}{!}{
\begin{tabular}{l c c c c c}
\toprule
model   & size & N@1   & N@5   & N@10  & N@50  \\
\midrule
Qwen2.5-VL  & 32B & 78.22 & 79.87 & 82.58 & 84.88 \\
Qwen2.5-VL  & 7B  & 74.16 & 77.26 & 80.26 & 82.84 \\
internVL & 8b  & \underline{80.28} & \underline{81.95} & \underline{84.66} & \underline{86.61} \\
Qwen3-VL    & 8B  & 72.45 & 75.48 & 78.32 & 81.44 \\
Qwen3-VL    & 4B  & 69.92 & 73.39 & 76.81 & 79.95 \\
Qwen3-VL-Reranker-8B   &8B  & 78.69 & 80.79 & 83.51 & 85.57 \\
lychee-reranker-mm & 8B& \textbf{92.35} & \textbf{93.41} & \textbf{94.42} & \textbf{94.86} \\
\bottomrule
\end{tabular}
}
\caption{NDCG@K for pointwise rerankers on MCMR evaluated on the LLaVE-7B top-50 candidate pool.}
\label{tab:reranking}
\end{table}


\section{Analysis}
\label{sec:analysis}

\begin{figure*}[ht]
\centering
\includegraphics[width=\linewidth]{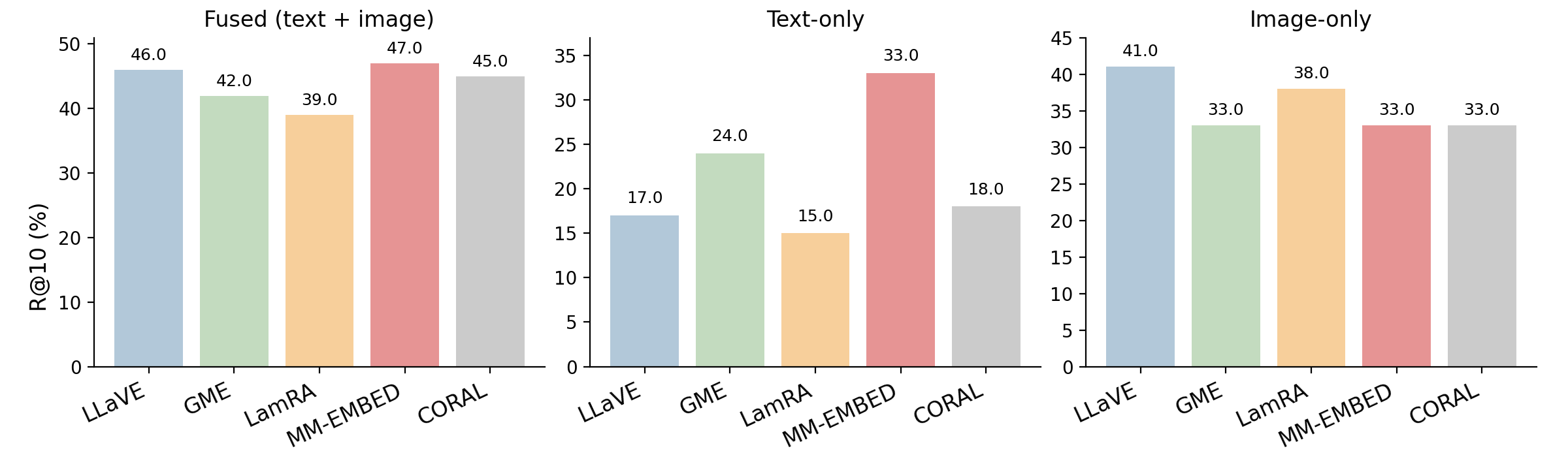}
\caption{Retrieval performance (Recall@10) under different query regimes. The complete results are provided in the supplementary material.}
\label{fig:r10_vs_q}
\end{figure*}

\subsection{Candidate-Side Modality Analysis}
\paragraph{Modality Contribution and Asymmetry.}
The modality ablation results in Tab.~\ref{tab:MCMR-bytype-subtables} show clear asymmetries in modality dependence. Under image-only candidates, GME and LamRA remain close to their fused performance at R@10, whereas LLaVE almost collapses once text is removed, with R@1 dropping from 24.99 to 0.90. Switching to text-only candidates also hurts all models relative to the fused setting, with MM-EMBED and CORAL exhibiting somewhat smaller drops at R@10.

Overall, text-only candidates perform far worse than fused candidates at R@10, and for four of the five shared models they also trail image-only candidates, indicating that visual cues are more discriminative than textual metadata alone on MCMR. Yet fused queries still outperform image-only by roughly 4–8 points at R@10, showing that textual metadata adds complementary constraints beyond visual evidence. This matches the dataset design, where each item includes at least one image-only and one text-only attribute, making purely unimodal reasoning incomplete.

\paragraph{Model-Level Variance and Design Implications.}
Modality removal amplifies cross-model variance. GME remains stable, whereas LLaVE and CORAL show strong early-rank degradation, suggesting a heavy reliance on textual priors. MM-EMBED, in contrast, appears more robust. These observations underscore a key limitation of current retrieval models: achieving strong global semantic similarity under fused inputs does not guarantee robustness when individual modalities weaken. Future designs may need to keep query conditions explicit and check each one separately, while also reducing redundant overlap between modalities to maintain performance.

\subsection{Query-Side Compositional Effects}
\paragraph{Query-Side Modality Ablation.}
To analyze modality contributions from the query side, we construct controlled query variants on a 100-query subset. We start from the fused natural-language queries, which are generated jointly from each product’s image\_feature list and text\_desc metadata. For the text-only condition, we use the ChatGPT-4o to compare each fused query against the image\_feature list and delete all phrases that correspond to visual attributes, yielding a version that retains only metadata-derived constraints. Symmetrically, for the image-only condition, we compare the fused query with the text\_desc and remove phrases grounded in textual metadata, preserving only visually grounded attributes. All resulting query triples, fused, text-only, and image-only, are manually checked for faithfulness, while the candidate pool is kept fixed with fused image–text metadata across all settings. 

Fig.~\ref{fig:r10_vs_q} summarizes retrieval performance at R@10 for three query regimes: fused text+image queries, text-only queries, and image-only queries.\footnote{Complete metrics for all models and query regimes are provided in the supplementary material.} Removing image-derived constraints in text-only queries causes severe early-rank degradation for every model: R@10 drops by around 20–30 points compared to the fused setting. In contrast, removing text-derived constraints in image-only queries leads to much smaller losses, and in several cases the image-only scores remain close to, or within roughly 5–10 points of, fused performance. The ranking of systems also shifts across regimes: MM-EMBED and LLaVE are strongest under fused queries, CORAL becomes the top performer in the text-only setting, and LLaVE clearly dominates when only visual evidence is available. Together, these patterns show that all systems benefit most from having both modalities present, but rely much more heavily on image cues than on textual metadata when forced into a single-modality regime.

\begin{figure}[ht]
\centering
\includegraphics[width=\linewidth]{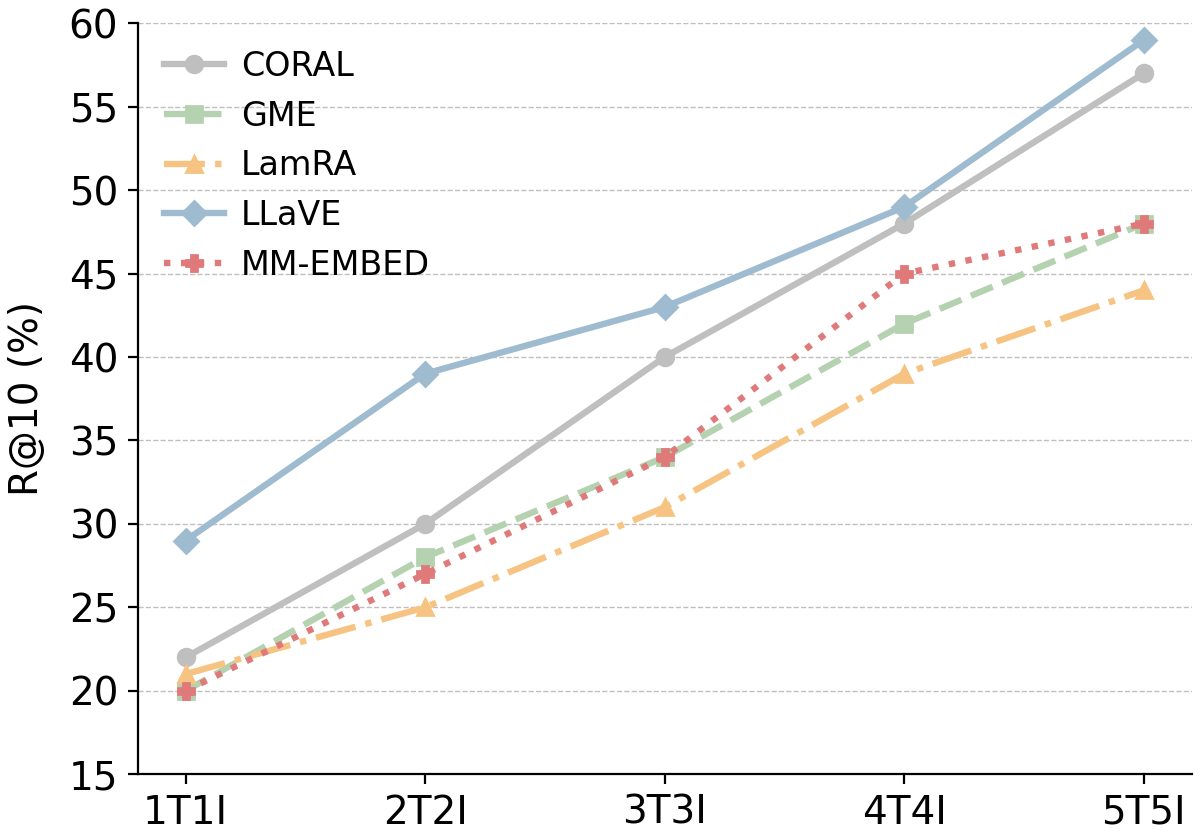}
\caption{Recall@10 under varying numbers of compositional constraints ($k_T = k_I \in {1,2,3,4,5}$). Candidates are fixed with fused image–text metadata.}
\label{fig:r10_vs_k}
\end{figure}

\paragraph{Effect of Query Constraint Count.}

We further vary the number of compositional constraints in the query. 
Let $k_T$ and $k_I$ denote the numbers of text-derived and image-derived constraints, respectively. 
We consider configurations where $k_T = k_I \in \{1,2,3,4,5\}$, using 1T+1I as a low-constraint baseline and 5T+5I as the most constrained setting, while fixing the candidate pool with fused image--text metadata and keeping the evaluation protocol unchanged. 
For each value of $k_T = k_I$, we use ChatGPT-4o to select $k$ visually grounded attributes from the image\_feature list and $k$ metadata-derived attributes from the text\_desc, and compose them into a single natural-language query. 
All resulting queries are manually checked for faithfulness and fluency, and all queries are encoded in text.

Fig.~\ref{fig:r10_vs_k} shows Recall@10 under varying numbers of compositional constraints per query, with equal counts of text-derived and image-derived constraints $k_T = k_I$ ranging from 1T+1I to 5T+5I, and with candidates fixed using fused image–text metadata.\footnote{A full table with Recall, MRR, and NDCG metrics for all models and constraint settings is provided in the supplementary material.} Across all models, performance consistently improves as more constraints are added, although the increase from 4T+4I to 5T+5I is smaller than the increase from 2T+2I to 3T+3I, indicating diminishing returns at higher counts.

\subsection{Pointwise Reranking Performance}

As shown in Tab.~\ref{tab:reranking}, all pointwise rerankers achieve strong early-rank accuracy, with NDCG@1 in the 70–80 range; the best model, InternVL-8B, reaches 80.3. When an MLLM compares a query and a candidate one-to-one, it can reliably identify the positive instance through explicit cross-modal grounding and condition verification. This stands in sharp contrast to the first-stage retrievers, whose best NDCG@1 is only 26.57 (CORAL), revealing a substantial performance gap. 
The substantial gap between reranking and first-stage retrieval indicates that current multimodal retrievers still struggle with fine-grained, multi-condition matching. Global embedding models emphasize holistic semantic similarity but often fail to verify whether all query conditions are simultaneously satisfied. In contrast, rerankers evaluate each query–candidate pair independently, enabling detailed cross-modal grounding and explicit constraint checking. The trade-off is computational efficiency: pointwise reranking is expensive and does not scale to large retrieval corpora.
These findings highlight a central challenge for future systems—integrating fine-grained multi-condition multimodal fusion into retrieval models while maintaining the scalability demanded by real-world applications.

\section{Conclusion}
\label{sec:conclusion}

In this work, we introduced MCMR (Multi-Conditional Multimodal Retrieval), a large-scale benchmark for evaluating fine-grained, multi-condition retrieval under natural-language queries. By explicitly disentangling visual and textual constraints, MCMR enables systematic assessment of models’ ability to reason over multiple interdependent conditions across modalities. Through comprehensive experiments on diverse multimodal retrievers and MLLM-based rerankers, we revealed consistent modality asymmetries—visual features dominate early-rank precision, while textual metadata stabilizes long-tail ordering. Moreover, MLLM-based pointwise rerankers substantially enhance fine-grained matching by explicitly verifying query–candidate consistency, though at a higher computational cost.
These findings expose a fundamental limitation in current multimodal retrieval systems: despite strong global semantic alignment, they struggle to integrate compositional, condition-aware reasoning at scale. We hope MCMR will serve as a diagnostic and challenging benchmark to inspire future research toward retrievers that jointly balance scalability, multimodal grounding, and compositional understanding.

{
    \small
    \bibliographystyle{ieeenat_fullname}
    \bibliography{main}

\begin{thebibliography}{49}
\providecommand{\natexlab}[1]{#1}
\providecommand{\url}[1]{\texttt{#1}}
\expandafter\ifx\csname urlstyle\endcsname\relax
  \providecommand{\doi}[1]{doi: #1}\else
  \providecommand{\doi}{doi: \begingroup \urlstyle{rm}\Url}\fi

\bibitem[Abootorabi et~al.(2025)Abootorabi, Zobeiri, Dehghani, Mohammadkhani, Mohammadi, Ghahroodi, Baghshah, and Asgari]{abootorabi2025ask}
Mohammad~Mahdi Abootorabi, Amirhosein Zobeiri, Mahdi Dehghani, Mohammadali Mohammadkhani, Bardia Mohammadi, Omid Ghahroodi, Mahdieh~Soleymani Baghshah, and Ehsaneddin Asgari.
\newblock Ask in any modality: A comprehensive survey on multimodal retrieval-augmented generation.
\newblock In \emph{Findings of the Association for Computational Linguistics: ACL 2025}, pages 16776--16809, Vienna, Austria, 2025. Association for Computational Linguistics.

\bibitem[Bai et~al.(2025)Bai, Chen, Liu, Wang, Ge, Song, Dang, Wang, Wang, Tang, Zhong, Zhu, Yang, Li, Wan, Wang, Ding, Fu, Xu, Ye, Zhang, Xie, Cheng, Zhang, Yang, Xu, and Lin]{Qwen2.5-VL}
Shuai Bai, Keqin Chen, Xuejing Liu, Jialin Wang, Wenbin Ge, Sibo Song, Kai Dang, Peng Wang, Shijie Wang, Jun Tang, Humen Zhong, Yuanzhi Zhu, Mingkun Yang, Zhaohai Li, Jianqiang Wan, Pengfei Wang, Wei Ding, Zheren Fu, Yiheng Xu, Jiabo Ye, Xi Zhang, Tianbao Xie, Zesen Cheng, Hang Zhang, Zhibo Yang, Haiyang Xu, and Junyang Lin.
\newblock Qwen2.5-vl technical report.
\newblock \emph{arXiv preprint arXiv:2502.13923}, 2025.

\bibitem[Chang et~al.(2022)Chang, Cao, Narang, Gao, Suzuki, and Bisk]{DBLP:conf/cvpr/ChangCNGSB22}
Yingshan Chang, Guihong Cao, Mridu Narang, Jianfeng Gao, Hisami Suzuki, and Yonatan Bisk.
\newblock Webqa: Multihop and multimodal {QA}.
\newblock In \emph{{IEEE/CVF} Conference on Computer Vision and Pattern Recognition}, pages 16474--16483, 2022.

\bibitem[Chen et~al.(2023)Chen, Hu, Luan, Sun, Changpinyo, Ritter, and Chang]{DBLP:conf/emnlp/ChenHLSCRC23}
Yang Chen, Hexiang Hu, Yi Luan, Haitian Sun, Soravit Changpinyo, Alan Ritter, and Ming-Wei Chang.
\newblock Can pre-trained vision and language models answer visual information-seeking questions?
\newblock In \emph{Proceedings of the 2023 Conference on Empirical Methods in Natural Language Processing}, pages 14948--14968, Singapore, 2023. Association for Computational Linguistics.

\bibitem[Chen et~al.(2024)Chen, Wu, Wang, Su, Chen, Xing, Zhong, Zhang, Zhu, Lu, et~al.]{chen2024internvl}
Zhe Chen, Jiannan Wu, Wenhai Wang, Weijie Su, Guo Chen, Sen Xing, Muyan Zhong, Qinglong Zhang, Xizhou Zhu, Lewei Lu, et~al.
\newblock Internvl: Scaling up vision foundation models and aligning for generic visual-linguistic tasks.
\newblock In \emph{Proceedings of the IEEE/CVF Conference on Computer Vision and Pattern Recognition}, pages 24185--24198, 2024.

\bibitem[Chow et~al.(2025{\natexlab{a}})Chow, Gao, Li, Wang, Xu, Song, Kong, Zhou, Zeng, Cai, Jiang, Xu, Zhang, Qiu, Li, Yang, Tang, and Li]{chow-etal-2025-merit-coral}
Wei Chow, Yuan Gao, Linfeng Li, Xian Wang, Qi Xu, Hang Song, Lingdong Kong, Ran Zhou, Yi Zeng, Yidong Cai, Botian Jiang, Shilin Xu, Jiajun Zhang, Minghui Qiu, Xiangtai Li, Tianshu Yang, Siliang Tang, and Juncheng Li.
\newblock Merit: Multilingual semantic retrieval with interleaved multi-condition query.
\newblock \emph{arXiv preprint arXiv:2506.03144}, 2025{\natexlab{a}}.
\newblock Introduces the \textsc{Coral} fine-tuning framework for multimodal retrieval.

\bibitem[Chow et~al.(2025{\natexlab{b}})Chow, Gao, Li, Wang, Xu, Song, Kong, Zhou, Zeng, Cai, Jiang, Xu, Zhang, Qiu, Li, Yang, Tang, and Li]{chow2025merit}
Wei Chow, Yuan Gao, Linfeng Li, Xian Wang, Qi Xu, Hang Song, Lingdong Kong, Ran Zhou, Yi Zeng, Yidong Cai, Botian Jiang, Shilin Xu, Jiajun Zhang, Minghui Qiu, Xiangtai Li, Tianshu Yang, Siliang Tang, and Juncheng Li.
\newblock Merit: Multilingual semantic retrieval with interleaved multi-condition query, 2025{\natexlab{b}}.

\bibitem[Dai et~al.(2025)Dai, Zhang, Li, Zhang, Long, Xie, Zhang, Li, and Zhang]{lychee}
Ziqi Dai, Xin Zhang, Mingxin Li, Yanzhao Zhang, Dingkun Long, Pengjun Xie, Meishan Zhang, Wenjie Li, and Min Zhang.
\newblock Supervised fine-tuning or contrastive learning? towards better multimodal llm reranking, 2025.

\bibitem[Faysse et~al.(2025)Faysse, Sibille, Wu, Omrani, Viaud, Hudelot, and Colombo]{faysse2024colpali}
Manuel Faysse, Hugues Sibille, Tony Wu, Bilel Omrani, Gautier Viaud, C{\'{e}}line Hudelot, and Pierre Colombo.
\newblock Colpali: Efficient document retrieval with vision language models.
\newblock In \emph{The Thirteenth International Conference on Learning Representations}, 2025.

\bibitem[Han et~al.(2017)Han, Wu, Huang, Zhang, Zhu, Li, Zhao, and Davis]{DBLP:conf/iccv/HanWHZZLZD17}
Xintong Han, Zuxuan Wu, Phoenix~X. Huang, Xiao Zhang, Menglong Zhu, Yuan Li, Yang Zhao, and Larry~S. Davis.
\newblock Automatic spatially-aware fashion concept discovery.
\newblock In \emph{{IEEE} International Conference on Computer Vision, {ICCV} 2017}, pages 1472--1480, Venice, Italy, 2017. {IEEE} Computer Society.

\bibitem[Han et~al.(2025)Han, Chen, Lin, Gao, Ren, Zhu, Ye, Wu, Xie, Gan, Wei, Xu, Wang, Zheng, Lin, Wu, and Ge]{han2025lemurlargescaleendtoend}
Xintian Han, Honggang Chen, Quan Lin, Jingyue Gao, Xiangyuan Ren, Lifei Zhu, Zhisheng Ye, Shikang Wu, XiongHang Xie, Xiaochu Gan, Bingzheng Wei, Peng Xu, Zhe Wang, Yuchao Zheng, Jingjian Lin, Di Wu, and Junfeng Ge.
\newblock Lemur: Large scale end-to-end multimodal recommendation, 2025.

\bibitem[Hou et~al.(2024)Hou, Li, He, Yan, Chen, and McAuley]{hou2024bridging}
Yupeng Hou, Jiacheng Li, Zhankui He, An Yan, Xiusi Chen, and Julian McAuley.
\newblock Bridging language and items for retrieval and recommendation.
\newblock \emph{arXiv preprint arXiv:2403.03952}, 2024.

\bibitem[Hu et~al.(2023)Hu, Luan, Chen, Khandelwal, Joshi, Lee, Toutanova, and Chang]{DBLP:conf/iccv/HuLCKJLTC23}
Hexiang Hu, Yi Luan, Yang Chen, Urvashi Khandelwal, Mandar Joshi, Kenton Lee, Kristina Toutanova, and Ming{-}Wei Chang.
\newblock Open-domain visual entity recognition: Towards recognizing millions of wikipedia entities.
\newblock In \emph{{IEEE/CVF} International Conference on Computer Vision, {ICCV} 2023}, pages 12031--12041, Paris, France, 2023. {IEEE}.

\bibitem[Jia et~al.(2021)Jia, Yang, Xia, Chen, Parekh, Pham, Le, Sung, Li, and Duerig]{jia2021scaling}
Chao Jia, Yinfei Yang, Ye Xia, Yi-Ting Chen, Zarana Parekh, Hieu Pham, Quoc Le, Yun-Hsuan Sung, Zhen Li, and Tom Duerig.
\newblock Scaling up visual and vision-language representation learning with noisy text supervision.
\newblock In \emph{International conference on machine learning}, pages 4904--4916. PMLR, 2021.

\bibitem[Jia et~al.(2025)Jia, Li, Yue, Li, Nie, Zou, and Chen]{jia2025visualwebinstruct}
Yiming Jia, Jiachen Li, Xiang Yue, Bo Li, Ping Nie, Kai Zou, and Wenhu Chen.
\newblock {V}isual{W}eb{I}nstruct: Scaling up multimodal instruction data through web search.
\newblock In \emph{Proceedings of the 2025 Conference on Empirical Methods in Natural Language Processing}, pages 1373--1393, Suzhou, China, 2025. Association for Computational Linguistics.

\bibitem[Jiang et~al.(2024{\natexlab{a}})Jiang, Song, Zhang, Huang, Deng, Sun, Zhang, Wang, and Zhuang]{jiang2024e5v}
Ting Jiang, Minghui Song, Zihan Zhang, Haizhen Huang, Weiwei Deng, Feng Sun, Qi Zhang, Deqing Wang, and Fuzhen Zhuang.
\newblock {E5-V:} universal embeddings with multimodal large language models.
\newblock \emph{CoRR}, abs/2407.12580, 2024{\natexlab{a}}.

\bibitem[Jiang et~al.(2024{\natexlab{b}})Jiang, Meng, Yang, Yavuz, Zhou, and Chen]{jiangvlm2vec}
Ziyan Jiang, Rui Meng, Xinyi Yang, Semih Yavuz, Yingbo Zhou, and Wenhu Chen.
\newblock Vlm2vec: Training vision-language models for massive multimodal embedding tasks.
\newblock In \emph{The Thirteenth International Conference on Learning Representations}, 2024{\natexlab{b}}.

\bibitem[Lan et~al.(2025)Lan, Niu, Meng, Zhou, and Su]{lan-etal-2025-llave}
Zhibin Lan, Liqiang Niu, Fandong Meng, Jie Zhou, and Jinsong Su.
\newblock {LL}a{VE}: Large language and vision embedding models with hardness-weighted contrastive learning.
\newblock In \emph{Findings of the Association for Computational Linguistics: EMNLP 2025}, pages 13721--13735, Suzhou, China, 2025. Association for Computational Linguistics.

\bibitem[Li et~al.(2022)Li, Li, Xiong, and Hoi]{li2022blip}
Junnan Li, Dongxu Li, Caiming Xiong, and Steven Hoi.
\newblock Blip: Bootstrapping language-image pre-training for unified vision-language understanding and generation.
\newblock In \emph{International conference on machine learning}, pages 12888--12900. PMLR, 2022.

\bibitem[Li et~al.(2026)Li, Zhang, Long, Chen, Song, Bai, Yang, Xie, Yang, Liu, Zhou, and Lin]{qwen3vlembedding}
Mingxin Li, Yanzhao Zhang, Dingkun Long, Keqin Chen, Sibo Song, Shuai Bai, Zhibo Yang, Pengjun Xie, An Yang, Dayiheng Liu, Jingren Zhou, and Junyang Lin.
\newblock Qwen3-vl-embedding and qwen3-vl-reranker: A unified framework for state-of-the-art multimodal retrieval and ranking.
\newblock \emph{arXiv}, 2026.

\bibitem[Lin et~al.(2024{\natexlab{a}})Lin, Lee, Shoeybi, Lin, Catanzaro, and Ping]{mm-embed}
Sheng-Chieh Lin, Chankyu Lee, Mohammad Shoeybi, Jimmy Lin, Bryan Catanzaro, and Wei Ping.
\newblock Mm-embed: Universal multimodal retrieval with multimodal llms.
\newblock In \emph{The Thirteenth International Conference on Learning Representations}, 2024{\natexlab{a}}.

\bibitem[Lin et~al.(2014)Lin, Maire, Belongie, Hays, Perona, Ramanan, Doll{\'{a}}r, and Zitnick]{lin2014microsoft}
Tsung{-}Yi Lin, Michael Maire, Serge~J. Belongie, James Hays, Pietro Perona, Deva Ramanan, Piotr Doll{\'{a}}r, and C.~Lawrence Zitnick.
\newblock Microsoft {COCO:} common objects in context.
\newblock In \emph{13th European Conference on Computer Vision, {ECCV} 2014}, pages 740--755, Zurich, Switzerland, 2014. Springer.

\bibitem[Lin et~al.(2024{\natexlab{b}})Lin, Mei, Chen, and Byrne]{lin-etal-2024-preflmr}
Weizhe Lin, Jingbiao Mei, Jinghong Chen, and Bill Byrne.
\newblock {P}re{FLMR}: Scaling up fine-grained late-interaction multi-modal retrievers.
\newblock In \emph{Proceedings of the 62nd Annual Meeting of the Association for Computational Linguistics (Volume 1: Long Papers)}, pages 5294--5316, Bangkok, Thailand, 2024{\natexlab{b}}. Association for Computational Linguistics.

\bibitem[Liu et~al.(2021{\natexlab{a}})Liu, Wang, Wang, and Ordonez]{DBLP:conf/emnlp/LiuWWO21}
Fuxiao Liu, Yinghan Wang, Tianlu Wang, and Vicente Ordonez.
\newblock Visual news: Benchmark and challenges in news image captioning.
\newblock In \emph{Proceedings of the 2021 Conference on Empirical Methods in Natural Language Processing}, pages 6761--6771, Online and Punta Cana, Dominican Republic, 2021{\natexlab{a}}. Association for Computational Linguistics.

\bibitem[Liu et~al.(2024)Liu, Hu, Xiao, Zhao, Gao, Wang, Li, and Tang]{multimodalrecommender}
Qidong Liu, Jiaxi Hu, Yutian Xiao, Xiangyu Zhao, Jingtong Gao, Wanyu Wang, Qing Li, and Jiliang Tang.
\newblock Multimodal recommender systems: A survey.
\newblock \emph{ACM Comput. Surv.}, 57\penalty0 (2), 2024.

\bibitem[Liu et~al.(2023{\natexlab{a}})Liu, Feng, Fu, Chen, and Wang]{DBLP:conf/emnlp/LiuFFCW23}
Siqi Liu, Weixi Feng, Tsu-Jui Fu, Wenhu Chen, and William Wang.
\newblock {EDIS}: Entity-driven image search over multimodal web content.
\newblock In \emph{Proceedings of the 2023 Conference on Empirical Methods in Natural Language Processing}, pages 4877--4894, Singapore, 2023{\natexlab{a}}. Association for Computational Linguistics.

\bibitem[Liu et~al.(2025)Liu, Chen, Cai, Jiang, Hu, Yao, Wang, and Xie]{liu-etal-2025-lamra}
Yikun Liu, Pingan Chen, Jiayin Cai, Xiaolong Jiang, Yao Hu, Jiangchao Yao, Yanfeng Wang, and Weidi Xie.
\newblock Lamra: Large multimodal model as your advanced retrieval assistant.
\newblock In \emph{Proceedings of the IEEE/CVF Conference on Computer Vision and Pattern Recognition (CVPR)}, pages 4015--4025. IEEE/CVF, 2025.

\bibitem[Liu et~al.(2021{\natexlab{b}})Liu, Opazo, Teney, and Gould]{Liu2021ImageRO}
Zheyuan Liu, Cristian~Rodriguez Opazo, Damien Teney, and Stephen Gould.
\newblock Image retrieval on real-life images with pre-trained vision-and-language models.
\newblock In \emph{2021 {IEEE/CVF} International Conference on Computer Vision, {ICCV} 2021}, pages 2105--2114, Montreal, Canada, 2021{\natexlab{b}}. {IEEE}.

\bibitem[Liu et~al.(2023{\natexlab{b}})Liu, Sun, Hong, Teney, and Gould]{liu2023bidirectionaltrainingcomposedimage}
Zheyuan Liu, Weixuan Sun, Yicong Hong, Damien Teney, and Stephen Gould.
\newblock Bi-directional training for composed image retrieval via text prompt learning, 2023{\natexlab{b}}.

\bibitem[Liu et~al.(2023{\natexlab{c}})Liu, Xiong, Lv, Liu, and Yu]{liu2023universal}
Zhenghao Liu, Chenyan Xiong, Yuanhuiyi Lv, Zhiyuan Liu, and Ge Yu.
\newblock Universal vision-language dense retrieval: Learning a unified representation space for multi-modal retrieval.
\newblock In \emph{The Eleventh International Conference on Learning Representations}, 2023{\natexlab{c}}.

\bibitem[Lu et~al.(2025{\natexlab{a}})Lu, Huang, Meng, Jin, Zeng, and Shen]{lu2025rethinking}
Xuan Lu, Haohang Huang, Rui Meng, Yaohui Jin, Wenjun Zeng, and Xiaoyu Shen.
\newblock Rethinking reasoning in document ranking: Why chain-of-thought falls short.
\newblock \emph{arXiv preprint arXiv:2510.08985}, 2025{\natexlab{a}}.

\bibitem[Lu et~al.(2025{\natexlab{b}})Lu, Liu, Yin, Li, Chen, Su, Jin, Zeng, and Shen]{lumulticonir}
Xuan Lu, Sifan Liu, Bochao Yin, Yongqi Li, Xinghao Chen, Hui Su, Yaohui Jin, Wenjun Zeng, and Xiaoyu Shen.
\newblock {M}ulti{C}on{IR}: Towards multi-condition information retrieval.
\newblock In \emph{Findings of the Association for Computational Linguistics: EMNLP 2025}, pages 13471--13494, Suzhou, China, 2025{\natexlab{b}}. Association for Computational Linguistics.

\bibitem[Lu et~al.(2026)Lu, Huang, Meng, Jin, Zeng, and Shen]{lu2026tools}
Xuan Lu, Haohang Huang, Rui Meng, Yaohui Jin, Wenjun Zeng, and Xiaoyu Shen.
\newblock Tools are under-documented: Simple document expansion boosts tool retrieval.
\newblock In \emph{The Fourteenth International Conference on Learning Representations}, 2026.

\bibitem[Luo et~al.(2023)Luo, Fang, Gokhale, Yang, and Baral]{luo-etal-2023-end}
Man Luo, Zhiyuan Fang, Tejas Gokhale, Yezhou Yang, and Chitta Baral.
\newblock End-to-end knowledge retrieval with multi-modal queries.
\newblock In \emph{Proceedings of the 61st Annual Meeting of the Association for Computational Linguistics (Volume 1: Long Papers)}, pages 8573--8589, Toronto, Canada, 2023. Association for Computational Linguistics.

\bibitem[Marino et~al.(2019)Marino, Rastegari, Farhadi, and Mottaghi]{DBLP:conf/cvpr/MarinoRFM19}
Kenneth Marino, Mohammad Rastegari, Ali Farhadi, and Roozbeh Mottaghi.
\newblock {OK-VQA:} {A} visual question answering benchmark requiring external knowledge.
\newblock In \emph{{IEEE} Conference on Computer Vision and Pattern Recognition, {CVPR} 2019}, pages 3195--3204, Long Beach, CA, USA, 2019. Computer Vision Foundation / {IEEE}.

\bibitem[Plummer et~al.(2015)Plummer, Wang, Cervantes, Caicedo, Hockenmaier, and Lazebnik]{DBLP:conf/iccv/PlummerWCCHL15}
Bryan~A. Plummer, Liwei Wang, Chris~M. Cervantes, Juan~C. Caicedo, Julia Hockenmaier, and Svetlana Lazebnik.
\newblock Flickr30k entities: Collecting region-to-phrase correspondences for richer image-to-sentence models.
\newblock In \emph{2015 {IEEE} International Conference on Computer Vision, {ICCV} 2015}, pages 2641--2649, Santiago, Chile, 2015. {IEEE} Computer Society.

\bibitem[Radford et~al.(2021)Radford, Kim, Hallacy, Ramesh, Goh, Agarwal, Sastry, Askell, Mishkin, Clark, et~al.]{2021clip}
Alec Radford, Jong~Wook Kim, Chris Hallacy, Aditya Ramesh, Gabriel Goh, Sandhini Agarwal, Girish Sastry, Amanda Askell, Pamela Mishkin, Jack Clark, et~al.
\newblock Learning transferable visual models from natural language supervision.
\newblock In \emph{International conference on machine learning}, pages 8748--8763. PmLR, 2021.

\bibitem[Shen et~al.(2022)Shen, Vakulenko, del Tredici, Barlacchi, Byrne, and de~Gispert]{shen2022lowresourcedenseretrievalopendomain}
Xiaoyu Shen, Svitlana Vakulenko, Marco del Tredici, Gianni Barlacchi, Bill Byrne, and Adrià de Gispert.
\newblock Low-resource dense retrieval for open-domain question answering: A comprehensive survey, 2022.

\bibitem[Shen et~al.(2023{\natexlab{a}})Shen, Asai, Byrne, and De~Gispert]{shen2023xpqa}
Xiaoyu Shen, Akari Asai, Bill Byrne, and Adria De~Gispert.
\newblock x{PQA}: Cross-lingual product question answering in 12 languages.
\newblock In \emph{Proceedings of the 61st Annual Meeting of the Association for Computational Linguistics (Volume 5: Industry Track)}, pages 103--115, Toronto, Canada, 2023{\natexlab{a}}. Association for Computational Linguistics.

\bibitem[Shen et~al.(2023{\natexlab{b}})Shen, Vakulenko, del Tredici, Barlacchi, Byrne, and de~Gispert]{shen2023neural}
Xiaoyu Shen, Svitlana Vakulenko, Marco del Tredici, Gianni Barlacchi, Bill Byrne, and Adria de Gispert.
\newblock Neural ranking with weak supervision for open-domain question answering : A survey.
\newblock In \emph{Findings of the Association for Computational Linguistics: EACL 2023}, pages 1736--1750, Dubrovnik, Croatia, 2023{\natexlab{b}}. Association for Computational Linguistics.

\bibitem[Wang et~al.(2024)Wang, Li, Zhu, Li, Zhang, and Shen]{wang2024crossmodalretrievalsystematicreview}
Tianshi Wang, Fengling Li, Lei Zhu, Jingjing Li, Zheng Zhang, and Heng~Tao Shen.
\newblock Cross-modal retrieval: A systematic review of methods and future directions, 2024.

\bibitem[Wei et~al.(2024)Wei, Chen, Chen, Hu, Zhang, Fu, Ritter, and Chen]{wei2023uniir}
Cong Wei, Yang Chen, Haonan Chen, Hexiang Hu, Ge Zhang, Jie Fu, Alan Ritter, and Wenhu Chen.
\newblock Uniir: Training and benchmarking universal multimodal information retrievers.
\newblock In \emph{18th European Conference on Computer Vision}, page 387–404, Milan, Italy, 2024. Springer-Verlag.

\bibitem[Wu et~al.(2021)Wu, Gao, Guo, Al{-}Halah, Rennie, Grauman, and Feris]{DBLP:conf/cvpr/WuGGARGF21}
Hui Wu, Yupeng Gao, Xiaoxiao Guo, Ziad Al{-}Halah, Steven Rennie, Kristen Grauman, and Rog{\'{e}}rio Feris.
\newblock Fashion {IQ:} {A} new dataset towards retrieving images by natural language feedback.
\newblock In \emph{{IEEE} Conference on Computer Vision and Pattern Recognition, {CVPR} 2021}, pages 11307--11317. Computer Vision Foundation / {IEEE}, 2021.

\bibitem[Zhang et~al.(2025{\natexlab{a}})Zhang, Han, and Han]{zhang2025research}
Bingbing Zhang, Yi Han, and Xiaofei Han.
\newblock Research on multi-modal retrieval system of e-commerce platform based on pre-training model.
\newblock \emph{Artificial Intelligence Technology Research}, 2\penalty0 (9), 2025{\natexlab{a}}.

\bibitem[Zhang et~al.(2025{\natexlab{b}})Zhang, Li, Zhang, Long, Li, Li, Xie, Zhang, Li, Zhang, et~al.]{zhangssrb}
Xin Zhang, Mingxin Li, Yanzhao Zhang, Dingkun Long, Yongqi Li, Yinghui Li, Pengjun Xie, Meishan Zhang, Wenjie Li, Min Zhang, et~al.
\newblock Ssrb: Direct natural language querying to massive heterogeneous semi-structured data.
\newblock In \emph{The Thirty-ninth Annual Conference on Neural Information Processing Systems Datasets and Benchmarks Track}, 2025{\natexlab{b}}.

\bibitem[Zhang et~al.(2025{\natexlab{c}})Zhang, Zhang, Xie, Li, Dai, Long, Xie, Zhang, Li, and Zhang]{zhang2025gme}
Xin Zhang, Yanzhao Zhang, Wen Xie, Mingxin Li, Ziqi Dai, Dingkun Long, Pengjun Xie, Meishan Zhang, Wenjie Li, and Min Zhang.
\newblock Bridging modalities: Improving universal multimodal retrieval by multimodal large language models.
\newblock In \emph{Proceedings of the Computer Vision and Pattern Recognition Conference}, pages 9274--9285, 2025{\natexlab{c}}.

\bibitem[Zhou et~al.(2024{\natexlab{a}})Zhou, Liu, Xiao, Zhao, and Xiong]{zhou-etal-2024-vista}
Junjie Zhou, Zheng Liu, Shitao Xiao, Bo Zhao, and Yongping Xiong.
\newblock {VISTA}: Visualized text embedding for universal multi-modal retrieval.
\newblock In \emph{Proceedings of the 62nd Annual Meeting of the Association for Computational Linguistics (Volume 1: Long Papers)}, pages 3185--3200, Bangkok, Thailand, 2024{\natexlab{a}}. Association for Computational Linguistics.

\bibitem[Zhou et~al.(2024{\natexlab{b}})Zhou, Zheng, Chen, Zheng, Su, Zhang, Meng, and Shen]{zhou2024beyond}
Jianqun Zhou, Yuanlei Zheng, Wei Chen, Qianqian Zheng, Hui Su, Wei Zhang, Rui Meng, and Xiaoyu Shen.
\newblock Beyond content relevance: Evaluating instruction following in retrieval models.
\newblock \emph{arXiv preprint arXiv:2410.23841}, 2024{\natexlab{b}}.

\bibitem[Zhou et~al.(2024{\natexlab{c}})Zhou, Mei, Li, Liu, Xiong, Liu, Gu, and Yu]{zhou-etal-2024-marvel}
Tianshuo Zhou, Sen Mei, Xinze Li, Zhenghao Liu, Chenyan Xiong, Zhiyuan Liu, Yu Gu, and Ge Yu.
\newblock {MARVEL}: Unlocking the multi-modal capability of dense retrieval via visual module plugin.
\newblock In \emph{Proceedings of the 62nd Annual Meeting of the Association for Computational Linguistics (Volume 1: Long Papers)}, pages 14608--14624, Bangkok, Thailand, 2024{\natexlab{c}}. Association for Computational Linguistics.

\end{thebibliography}
}

\clearpage
\setcounter{page}{1}
\maketitlesupplementary

\section{Details of Baselines}

For each model used in this paper, Tab.~\ref{tab:baselines} summarizes the parameter size, architecture, whether we use an explicit retrieval instruction, the maximum input context length, and the underlying backbone checkpoint. We group systems into retrievers and rerankers according to their role in our two-stage pipeline.

\begin{table*}[t]
\centering
\scriptsize
\setlength{\tabcolsep}{4.5pt}
\resizebox{\textwidth}{!}{%
\begin{tabular}{l c *{8}{c}}
\toprule
Model & Size & Architecture & Instruction & Max length & Backbone \\
\midrule
\multicolumn{6}{c}{Retriever} \\
\midrule
GME-Qwen2-VL-7B-Instruct   & 7B  & Decoder & Yes & 32K  & Qwen2-VL-7B-Instruct \\
LLaVE-7B                   & 7B  & Decoder & Yes & 32K  & LLaVA-OneVision-7B   \\
LamRA-Ret-Qwen2.5VL-7B     & 7B  & Decoder & Yes & 128K & Qwen2.5-VL-7B       \\
MM-Embed                   & 8B  & Decoder & Yes & 32K  & NV-Embed            \\
CORAL                      & 3B  & Decoder & No  & 128K & Qwen2.5-3B-Instruct \\
VLM2VEC                    & 4B  & Decoder & No  & 131K & Phi3.5V            \\
\midrule
\multicolumn{6}{c}{MLLM-as-Reranker} \\
\midrule
Qwen2.5-VL-7B-Instruct     & 7B   & Decoder & Yes & 128K & Qwen2.5-VL-7B-Instruct  \\
Qwen2.5-VL-32B-Instruct    & 32B  & Decoder & Yes & 128K & Qwen2.5-VL-32B-Instruct \\
InternVL3-8B-Instruct      & 32B  & Decoder & Yes & 32K  & InternVL3-8B-Instruct   \\
Qwen3-VL-4B-Instruct       & 4B   & Decoder & Yes & 262K & Qwen3-VL-4B-Instruct    \\
Qwen3-VL-8B-Instruct       & 8B   & Decoder & Yes & 262K & Qwen3-VL-8B-Instruct    \\
\bottomrule
\end{tabular}}%
\vspace{4.5pt}
\caption{Details of retriever and reranker models used in our experiments.}
\label{tab:baselines}
\end{table*}

Tab.~\ref{tab:baselines}: Details of retriever and reranker models used in experiments. Size denotes the number of parameters of each model. Architecture indicates the model type used in our setup (all models are operated in decoder-only mode). Instruction specifies whether we prepend a task-specific natural-language prefix when encoding queries (for example, retrieval prompts or templates that produce an \texttt{<emb>} token), rather than feeding bare queries. Max length denotes the maximum input length in tokens used for each model in our experiments, typically matching the backbone context window. Backbone identifies the pretrained checkpoint on which each retriever or reranker is built.

\section{Examples of MCMR}
Tab.~\ref{tab:mcmr_examples} presents examples from the five MCMR sub-datasets, covering \textit{upper clothing}, \textit{bottom clothing}, \textit{shoes}, \textit{jewelry}, and \textit{furniture}.

\begin{table*}[t]
\centering
\resizebox{\textwidth}{!}{
\begin{tabular}{L{2.2cm} L{6.2cm} L{6.2cm} C{3.5cm}}
\toprule
\textbf{Category} & \textbf{Query Text} & \textbf{Target Text} & \textbf{Target Image} \\
\midrule

Upper &
I’m looking for a men’s jacket in gray with a plaid pattern and green accents, size L. It should be waterproof with a hood and front pockets, made of durable nylon twill that needs hand washing. Prefer something released around 2013 and priced about \$200. &
"title": "Columbia Men's Whirlibird III Interchange Jacket",
"description": "Three jackets in one – a warm, repellent Omni-Heat liner; a waterproof-breathable and critically seam-sealed shell; and a combination of both – giving you ultimate versatility to stay warm, dry, protected and comfortable in fluctuating winter conditions.",
"price": 200.0, ... &
\includegraphics[width=2.8cm]{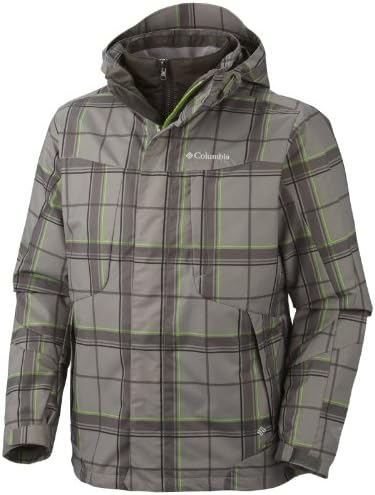} \\
\cmidrule(lr){1-4}\addlinespace[0.4em]

Bottom &
I’m looking for a pair of men’s brown cotton jeans with a slim, straight-leg fit. I’d like a flat front and zipper fly, made from 100\% cotton denim that’s soft, durable, and machine washable. Perfect for casual or work wear, ideally under \$30 and released around 2021. &
"title": "World of Leggings Plus Size Spandex Knee High Boy Shorts - Shop 16 Colors",
"description": "Our knee high and seamless plus size boy shorts are a one size seamless nylon spandex plus size leg piece that are a must for any woman's leggings wardrobe. They are made with high quality stitching and have fantastic stretch ... &
\includegraphics[width=2.6cm]{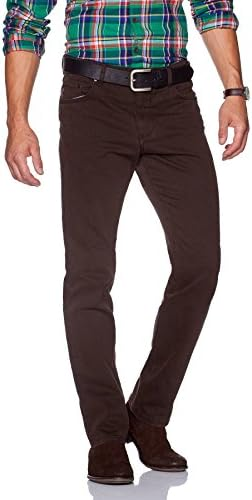} \\
\cmidrule(lr){1-4}\addlinespace[0.4em]

Shoes &
I’m looking for men’s brown leather high-top work boots with a lace-up closure and a cushioning OrthoLite footbed. They should have slip-resistant soles and meet ASTM EH safety standards for electrical hazard protection. I’d like a pair under \$260. &
"title": "Danner Men's Bull Run 8\" Work Boot",
"description": "Built in the USA with a dedication to quality that goes back to 1932, the Bull Run is a utilitarian work boot with a timeless design that stays in style when you punch out. The full-grain leather upper is the perfect blend of strong and soft. Combine that with the sturdy Danner Wedge outsole and you get all-day comfort that lasts.",
"price": 259.95,
"features": "100\% Leather", ... &
\includegraphics[width=2.8cm]{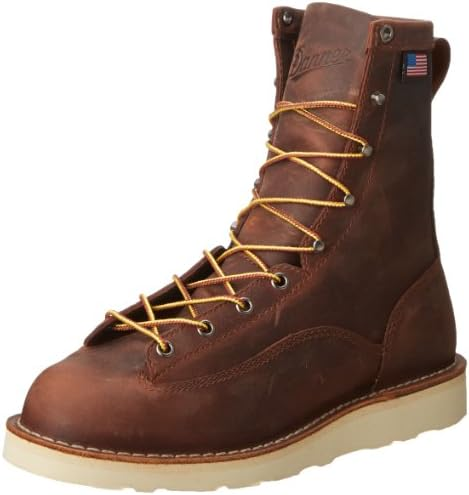} \\
\cmidrule(lr){1-4}\addlinespace[0.4em]

Jewelry &
I’m looking for a silver-tone bracelet with a Cuban link chain and slide clasp. Please show options in 14k gold over 925 sterling silver, about \$450, released around 2022. &
"title": "SAVEARTH DIAMONDS 5.80 Ct to 10.90 Ct Lab Created Moissanite Diamond 6MM Width Cuban Link Chain Necklace For Men In 14k Gold Over 925 Sterling Silver, 16\" to 30\" Length, Color: G-H, Clarity: VVS1",
"description": "Jewelry has the power to be this one little thing that can make you feel unique, ...", ... &
\includegraphics[width=2.8cm]{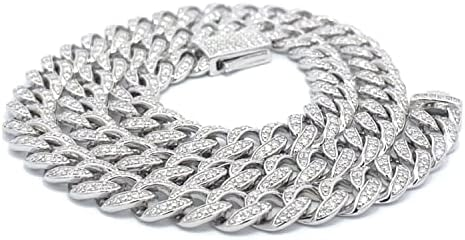} \\
\cmidrule(lr){1-4}\addlinespace[0.4em]

Furniture &
I’m looking for some cute rustic Halloween decorations for my home, i want a tiered-tray setup with small wooden signs and fall-themed accents, under \$20. &
"title": "CYNOSA Halloween Decorations Halloween Tiered Tray Decor Fall Decor Hocus Pocus I Smell Children Boo Wooden Signs and Orange Plaid Gnomes Plush Farmhouse Rustic Tiered Tray Decor for Home Table",
"description": "Package including: 1 x Black and Orange Check Plaid Gnome; 1 x Round Shape Sign \"I Smell Children\"; 1 x Square Shape Halloween Themed Sign \"October 31\"; 1 x Rectangle Shape Black and Orange Plaid Sign (Boo!); ... &
\includegraphics[width=2.8cm]{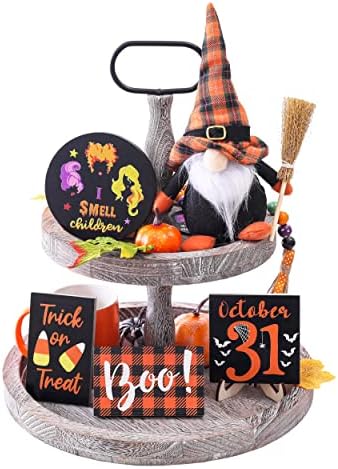} \\
\bottomrule
\end{tabular}
}
\vspace{0.8em}
\caption{Examples of retrieval samples from our MCMR benchmark.}
\label{tab:mcmr_examples}
\end{table*}

\section{Complete Results}

Tab.~\ref{tab:mcmr-results-ref-table} reports the full numerical results of our query-side modality ablation study. 
For each of the five retrievers evaluated in our main experiments, we provide detailed scores under three query regimes: (i) fused queries containing both image-derived and text-derived constraints, (ii) text-only queries obtained by removing all image-grounded constraints, and (iii) image-only queries obtained by removing all text-grounded constraints. 
Each subtable reports Recall@K, MRR, and NDCG@K across a broad range of cutoffs, enabling fine-grained inspection of early-rank precision as well as long-range ordering behavior.
These full results make clear how different models respond to the removal of cross-modal evidence. 

Tab.~\ref{tab:ablation-kTikI} reports the full numerical results for the query-side compositional constraint study described in §4.3. For each retriever (CORAL, GME, LamRA, LLaVE, and MM-Embed) and for each configuration of the text- and image-derived constraint counts, we list Recall@K, MRR, and NDCG@K in percentage form. We consider matched constraint settings with $k_T{=}k_I\in\{1,2,3,4,5\}$ while keeping the fused candidate pool and evaluation protocol identical to the main experiments. The main text focuses on the trends for $k_T{=}k_I\in\{2,3,4,5\}$; here we additionally include the $1\mathrm{T}{+}1\mathrm{I}$ configuration for completeness.

\begin{table*}[t]
\centering
\scriptsize
\setlength{\tabcolsep}{5pt}

\begin{subtable}{\linewidth}
\centering
\subcaption{Fused(text+image)}
\resizebox{\linewidth}{!}{
\begin{tabular}{l c r r r r r r r r r r r r}
\toprule
model & R@1 & R@5 & R@10 & R@50 & MRR & N@5 & N@10 & N@50 \\
\midrule
LLaVE        & 22.00 & 38.00 & 46.00 & 73.00 & 29.09 & 30.53 & 33.10 & 39.39 \\
GME-Qwen2VL  &  9.00 & 34.00 & 42.00 & 64.00 & 19.50 & 22.32 & 24.91 & 29.81 \\
LamRA-Qwen2.5VL & 13.00 & 31.00 & 39.00 & 63.00 & 21.14 & 22.73 & 25.43 & 30.67 \\
MM-EMBED     & 20.00 & 40.00 & 47.00 & 64.00 & 27.58 & 29.94 & 32.19 & 35.88 \\
CORAL        & 24.00 & 39.00 & 45.00 & 71.00 & 30.67 & 32.22 & 34.11 & 39.86 \\
\bottomrule
\end{tabular}}
\end{subtable}

\vspace{0.8em}

\begin{subtable}{\linewidth}
\centering
\subcaption{Text-only}
\resizebox{\linewidth}{!}{
\begin{tabular}{l c r r r r r r r r r r r r}
\toprule
model & R@1 & R@5 & R@10 & R@50 & MRR & N@5 & N@10 & N@50 \\
\midrule
LLaVE        &  2.00 & 12.00 & 17.00 & 30.00 &  6.54 &  7.40 &  9.02 & 11.98 \\
GME-Qwen2VL  &  6.00 & 16.00 & 24.00 & 45.00 & 11.09 & 11.65 & 14.08 & 18.43 \\
LamRA-Qwen2.5VL &  2.00 &  8.00 & 15.00 & 27.00 &  4.77 &  4.92 &  7.10 &  9.68 \\
MM-EMBED     & 12.00 & 29.00 & 33.00 & 47.00 & 18.72 & 20.90 & 22.18 & 25.47 \\
CORAL        &  9.00 & 16.00 & 18.00 & 30.00 & 12.00 & 12.82 & 13.45 & 16.05 \\
\bottomrule
\end{tabular}}
\end{subtable}

\vspace{0.8em}

\begin{subtable}{\linewidth}
\centering
\subcaption{Image-only}
\resizebox{\linewidth}{!}{
\begin{tabular}{l c r r r r r r r r r r r r}
\toprule
model & R@1 & R@5 & R@10 & R@50 & MRR & N@5 & N@10 & N@50 \\
\midrule
LLaVE        & 20.00 & 33.00 & 41.00 & 65.00 & 26.04 & 27.06 & 29.55 & 35.07 \\
GME-Qwen2VL  & 13.00 & 25.00 & 33.00 & 56.00 & 18.07 & 18.91 & 21.57 & 26.41 \\
LamRA-Qwen2.5VL & 12.00 & 30.00 & 38.00 & 60.00 & 19.11 & 20.94 & 23.57 & 28.42 \\
MM-EMBED     & 14.00 & 28.00 & 33.00 & 50.00 & 19.42 & 21.09 & 22.64 & 26.59 \\
CORAL        & 10.00 & 31.00 & 37.00 & 54.00 & 18.82 & 21.27 & 23.21 & 26.81 \\
\bottomrule
\end{tabular}}
\end{subtable}

\caption{Retrieval on MCMR under three query-visibility regimes (fused / text-only / image-only). Values are percentages.}
\label{tab:mcmr-results-ref-table}
\end{table*}

\begin{table*}[t]
\centering
\scriptsize
\setlength{\tabcolsep}{4.5pt}
\resizebox{\textwidth}{!}{%
\begin{tabular}{l c *{8}{c}}
\toprule
\multirow{2}{*}{model} & \multirow{2}{*}{$k_T{=}k_I$} & \multicolumn{4}{c}{Recall@K (\%)} & \multirow{2}{*}{MRR (\%)} & \multicolumn{3}{c}{NDCG@K (\%)} \\
\cmidrule(lr){3-6}\cmidrule(lr){8-10}
& & {1} & {5} & {10} & {50} & & {5} & {10} & {50} \\
\midrule
\multirow{5}{*}{CORAL}
& 1t1i & 9.00 & 15.00 & 22.00 & 41.00 & 12.38 & 12.30 & 14.61 & 18.84 \\
& 2t2i & 15.00 & 31.00 & 38.00 & 57.00 & 22.09 & 23.64 & 25.89 & 30.11 \\
& 3t3i & 21.00 & 36.00 & 40.00 & 62.00 & 27.25 & 29.06 & 30.34 & 35.10 \\
& 4t4i & 26.00 & 42.00 & 48.00 & 68.00 & 33.21 & 34.86 & 36.77 & 41.13 \\
& 5t5i & 33.00 & 50.00 & 57.00 & 79.00 & 40.19 & 41.94 & 44.19 & 48.93 \\
\cmidrule(lr){1-10}
\multirow{5}{*}{GME}
& 1t1i & 2.00 & 18.00 & 20.00 & 36.00 & 7.92 & 10.22 & 10.87 & 14.32 \\
& 2t2i & 10.00 & 23.00 & 28.00 & 46.00 & 15.75 & 17.03 & 18.70 & 22.67 \\
& 3t3i & 12.00 & 32.00 & 34.00 & 52.00 & 19.43 & 22.32 & 23.01 & 27.13 \\
& 4t4i & 18.00 & 39.00 & 42.00 & 59.00 & 25.78 & 28.73 & 29.73 & 33.55 \\
& 5t5i & 21.00 & 42.00 & 48.00 & 68.00 & 29.40 & 31.93 & 33.87 & 38.36 \\
\cmidrule(lr){1-10}
\multirow{5}{*}{LamRA}
& 1t1i & 5.00 & 12.00 & 21.00 & 42.00 & 8.51 & 8.49 & 11.38 & 15.94 \\
& 2t2i & 8.00 & 19.00 & 25.00 & 50.00 & 13.06 & 13.90 & 15.90 & 21.38 \\
& 3t3i & 8.00 & 24.00 & 31.00 & 53.00 & 14.78 & 16.31 & 18.67 & 23.80 \\
& 4t4i & 12.00 & 29.00 & 39.00 & 63.00 & 19.48 & 20.81 & 24.08 & 29.41 \\
& 5t5i & 19.00 & 35.00 & 44.00 & 70.00 & 26.66 & 27.84 & 30.79 & 36.52 \\
\cmidrule(lr){1-10}
\multirow{5}{*}{LLaVE}
& 1t1i & 5.00 & 23.00 & 29.00 & 50.00 & 12.55 & 14.57 & 16.49 & 21.14 \\
& 2t2i & 16.00 & 32.00 & 39.00 & 57.00 & 21.83 & 23.49 & 24.54 & 29.67 \\
& 3t3i & 19.00 & 37.00 & 43.00 & 68.00 & 25.98 & 28.06 & 30.04 & 35.45 \\
& 4t4i & 21.00 & 42.00 & 49.00 & 69.00 & 28.93 & 31.40 & 33.72 & 38.10 \\
& 5t5i & 27.00 & 50.00 & 59.00 & 72.00 & 35.89 & 38.50 & 41.35 & 44.28 \\
\cmidrule(lr){1-10}
\multirow{5}{*}{MM-EMBED}
& 1t1i & 8.00 & 18.00 & 20.00 & 40.00 & 12.06 & 13.36 & 13.98 & 18.57 \\
& 2t2i & 14.00 & 22.00 & 27.00 & 44.00 & 17.31 & 18.03 & 19.55 & 22.09 \\
& 3t3i & 19.00 & 29.00 & 34.00 & 49.00 & 22.96 & 23.96 & 25.55 & 28.80 \\
& 4t4i & 23.00 & 40.00 & 45.00 & 64.00 & 30.04 & 31.96 & 33.65 & 37.83 \\
& 5t5i & 30.00 & 43.00 & 48.00 & 68.00 & 36.30 & 37.43 & 39.15 & 43.67 \\
\bottomrule
\end{tabular}}%
\vspace{4pt}
\caption{Query-side modality ablation under compositional constraints ($k_T{=}k_I\in\{1,2,3,4,5\}$). Values are percentages.}
\label{tab:ablation-kTikI}
\end{table*}

\section{Prompts}
\label{sec:prompts}

\subsection{Prompts for Constructing MMR}


\paragraph{Step 1: Image-side Attribute Extraction}

Fig.~\ref{fig:image-attr-prompt} shows the prompt template used to generate fine-grained visual attribute descriptions from each product image. The model is instructed to output structured, image-grounded features only, without inferring hidden or functional properties.

\begin{figure*}[t]
\centering
\begin{minipage}{\textwidth}
\begin{promptbox}[Prompt for Image Attribute Extraction]
You are a meticulous vision annotator.

Task

	1.	Identify the product category you see in one or two lowercase words.
    
	2.	List short appearance descriptors capturing fine-grained, objective visual details of the product only.

Rules

	•	Output exactly one JSON array: the category first, then the descriptors.
    
	•	Aim for 4–10 descriptors; if fewer are certain, output only those (do not guess).
    
	•	Each descriptor $\leq$ 8 English words, all lowercase, American spelling.
    
	•	Describe only what is clearly visible; do not mention size, price, release date, performance or comfort claims.
    
	•	If a word implies function/technology (e.g., waterproof, breathable, thermal, cushioned), skip it unless the exact text is visibly printed.
    
	•	Focus on: colors, materials/textures, local patterns/graphics, construction/parts/edges, seams/overlays/stitching, closure/hardware (e.g., zipper, buckle, laces, clasp), shape/silhouette, logos/text when readable.
    
	•	If readable logos or text appear, include the exact lowercase text (no quotes). Do not invent brands or hallmarks.
    
	•	Exclude background or other items not part of the product (e.g., other garments, props, body parts).
    
	•	Skip any feature you are not 100\% certain is visible; do not infer hidden details.

Before you output, ensure coverage of at least 4 of these slots if visible: 
\{metal color{/}finish\}, \{stone color \& shape{/}cut\}, \{setting{/}arrangement\}, 
\{band{/}chain{/}bracelet details\}, \{closure{/}clasp\}, 
\{engraving{/}motif{/}overlays\}, \{logos{/}text{/}hallmarks\}.

Return only the JSON array (no prose).

\end{promptbox}
\end{minipage}
\caption{Prompt for Image Attribute Extraction used in image-side annotation.}
\label{fig:image-attr-prompt}
\end{figure*}



\paragraph{Step 2: Text-side Description Generation}

Fig.~\ref{fig:text-desc-prompt} illustrates the prompt design for generating textual product descriptions from metadata. 
It enforces strict separation from visual evidence and standardizes phrasing for key attributes such as price and release date.

\begin{figure*}[t]
\centering
\begin{minipage}{\textwidth}
\begin{promptbox}[Text-side Description Generation]
Task: Generate a concise English description for a product from its JSON metadata.

Inputs (use only text fields; the last array is a FORBIDDEN list, not content to describe):
- title: {title}
- description: {description}
- features: {features}
- price: {price}
- Date First Available: {date}
- forbidden\_visuals (from image\_feature; DO NOT mention or paraphrase): {image}

Hard rules

• Write 2--4 clear sentences, total 80--120 words, single paragraph.

• Use only these fields: title, description, features, price, Date First Available. Ignore images and any visual traits.

• Always mention price and first-availability time with strict phrasing:

  -- Price: strictly ``priced at \$X.XX''. If absent: ``price information not provided''.
  
  -- Date: strictly ``released <Month> <Day>, <Year>''. If absent: ``release date not provided''.
  
• Do not include any visual attributes (colors, patterns, graphics, logos, slogans, characters, shape, silhouette, finish, gloss, matte, style cues).

• Treat every item in forbidden\_visuals as banned; do not include those words or their paraphrases.

• Focus only on text-sourced facts: product type, use, materials, components, construction, closures, care, durability, explicit sizing, measurements, certifications, standards, packaging, warranty, origin, manufacturing, price, release date.

• Exclude shipping, service, and marketing slogans; compress long size tables into concise ranges when necessary.

• Output only the final paragraph, no notes or reasoning.

Quality gate (internal)

• Remove any overlap with forbidden\_visuals or residual visual wording.

• Enforce length 80--120 words; if short and facts exist, add non-visual facts from the text (e.g., care, sizing, certification, origin).

• If fields conflict, prefer ``features'' over ``description''. Never invent facts.

\medskip
\noindent{Input JSON template}
\begin{verbatim}
{
  "title": {title},
  "description": {description},
  "features": {features},
  "price": {price},
  "Date First Available": {date},
  "image_feature": {image}
}
\end{verbatim}
\end{promptbox}
\caption{Prompt template used for text-side description generation. }
\label{fig:text-desc-prompt}
\end{minipage}
\end{figure*}




\paragraph{Step 3: Universal Leakage Checker }

Fig.~\ref{fig:leakage-prompt} shows the verification prompt used to detect visual-to-text leakage after description generation.
It flags any exact or paraphrased overlap between \texttt{image\_feature} and \texttt{text\_desc}, as well as category mismatches.
This step guarantees that text-side descriptions remain strictly text-grounded before query synthesis.

\begin{figure*}[t]
\centering
\begin{minipage}{\textwidth}
\begin{promptbox}[Prompt for Cross-Modal Leakage Detection]
Return ONE valid JSON object ONLY, no prose.

\medskip
Task: Check if any content in image\_feature appears in text\_desc as an exact phrase or a clear paraphrase or synonym. Use case\-insensitive matching. Treat hyphen and space as equivalent. Handle simple singular or plural.

\medskip
Category: If the first image\_feature item is a generic category , ignore it for leakage counting, but set category\_conflict=true only if text\_desc clearly names a different category.

\medskip
Disambiguation: Do not count material specifications like 925 sterling silver'' as a match for the visual tone silver\-tone metal’’. Be conservative: mark paraphrase only when meaning is the same.

\medskip
Output JSON schema:\
{\
\quad “leak”: true, false,\
\quad “matches”: [{ “feature”: str, “type”: “exact, paraphrase”, “evidence”: str }],\
\quad “category\_conflict”: true, false\
}\

\medskip
Input:\
image\_feature: [\ldots]\
text\_desc: ``\ldots’’

\medskip
Return JSON now.
\end{promptbox}
\caption{Prompt template for cross-modal leakage detection.}
\label{fig:leakage-prompt}
\end{minipage}
\end{figure*}


\paragraph{Step 4: Query Generation }

Fig.~\ref{fig:query-gen-prompt} illustrates the prompt used to simulate realistic customer queries.
The model combines visual and textual attributes under strict composition rules, ensuring each query naturally expresses cross-modal information while maintaining linguistic diversity and authenticity.

\begin{figure*}[t]
\centering
\begin{minipage}{\textwidth}
\begin{promptbox}[Prompt for Query Generation]
You are a real shopper typing in the search box of a large e-commerce site.

Task

Write ONE fluent search query in a natural first-person voice. Use 2-3 sentences and keep the total length between 35-60 words.

Sources
\begin{itemize}
\item Description : {desc}
\item Image tags : {image}
\end{itemize}

Hard rules — all must be met
\begin{itemize}
\item Blend both sources -> Use exactly 2-3 visual tags and exactly 2-3 description facts, with no overlaps.
\begin{itemize}
\item If fewer than 3 description facts remain after removing overlaps, use all remaining and do not invent any.
\item If a clear pattern appears in Image tags, specify the base color and the pattern’s color, and use exactly one generic pattern term such as “floral print”, “graphic motif”, “striped pattern”. Otherwise mention only the base color.
\item Mention exactly one target size only if present, never list multiple sizes or ranges.
\end{itemize}
\item Price format  Add one approximate price using “about” or “under” only if a price appears in the sources, never use the exact price.
\item Release time format  Add one broad release time using a relative or approximate expression only if a date appears in the sources, never use exact dates.
\item Use first-person tone throughout.
\item No details beyond the two sources.
\item Output only the query text, no bullet points, quotes, or backslashes.
\end{itemize}

Materials phrasing
\begin{itemize}
\item When the Description lists precise fiber percentages, rewrite them into qualitative phrases with no digits and no percent sign, for example “cotton-rich”, “with some polyester”, “with a touch of elastane”, “all-cotton”, “polyester blend”, “nylon-rich”, “wool blend”. Each such materials phrase counts as one description fact.
\end{itemize}

Style hints
\begin{itemize}
\item Emphasize fabric, fit, closure, care before marketing fluff.
\item Vary phrasing so each query feels unique.
\item Keep adjectives objective, for example “lightweight cotton jersey”, not “super comfy”.
\end{itemize}

Input JSON

{{\
\quad “desc”: {desc},\

\quad “image\_feature”: {image}\
}}

Output

\textless query text only\textgreater
\end{promptbox}
\caption{Prompt template for query generation.}
\label{fig:query-gen-prompt}
\end{minipage}
\end{figure*}


\paragraph{Step 5: Query Check.}

\begin{figure*}[t]
    \centering
    \includegraphics[width=\textwidth]{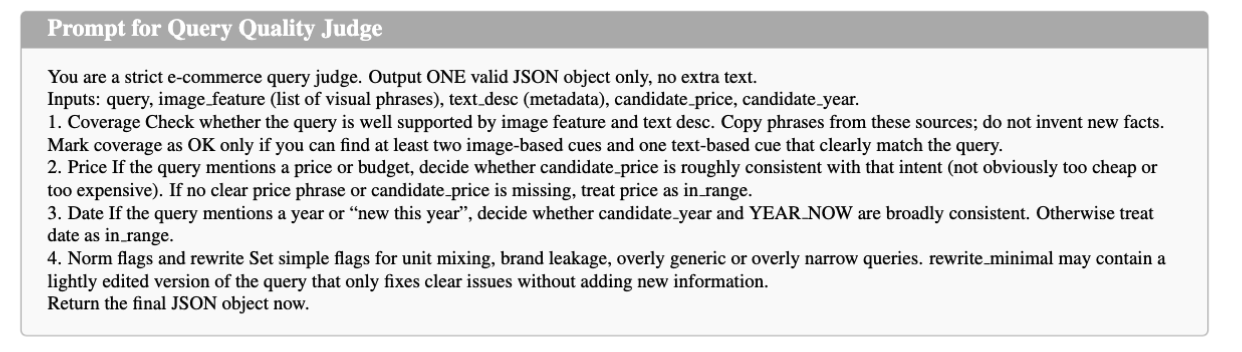}%
    \caption{Prompt for query quality judge.}
    \label{fig:query-verifier-prompt}
\end{figure*}

Fig.~\ref{fig:query-verifier-prompt} presents the verification prompt that filters low-quality or unfaithful queries.
The model checks for balanced modality coverage (at least two image-derived and one text-derived matches), valid price and date phrasing, and normalization issues such as unit inconsistencies or brand leakage.
Only queries passing all logical gates are retained for dataset construction.

\paragraph{Pointwise.}

\begin{figure*}[t]
    \centering
    \includegraphics[width=\textwidth]{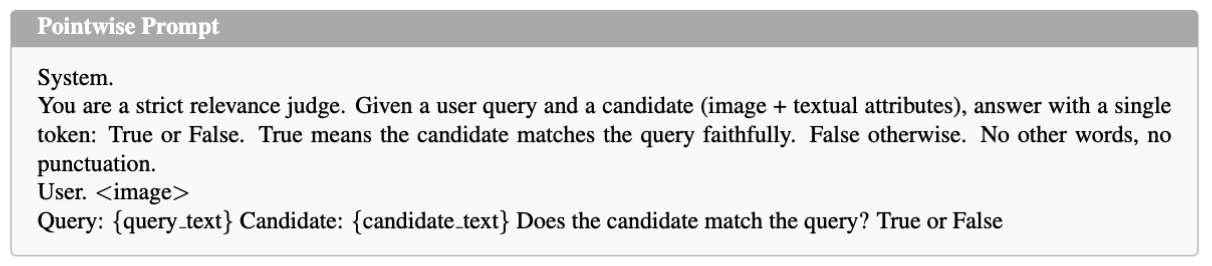}%
    \caption{Prompt used for pointwise relevance scoring.}
    \label{fig:point-prompt}
\end{figure*}

Fig.~\ref{fig:point-prompt} shows the pointwise prompt used to score query–candidate pairs during reranking.

\section{Human Evaluation Protocol}

\begin{table}[ht]
\centering
\small
\resizebox{\linewidth}{!}{
\begin{tabular}{lll}
\toprule
\textbf{Annotator} & \textbf{Background} & \textbf{English Proficiency} \\
\midrule
Annotator A &
B.A.\ in Linguistics &
IELTS~7.0 \\
Annotator B &
M.S.\ in Computer Science &
TOEFL~99 \\
\bottomrule
\end{tabular}
}
\caption{Background and qualifications of annotators involved in the human evaluation study.}
\end{table}

\begin{table}[ht]
\centering
\small
\resizebox{\linewidth}{!}{
\begin{tabular}{lcc}
\toprule
\textbf{Metric} & \textbf{Human} & \textbf{Generated}  \\
\midrule
Attribute correctness & 4.45 & 4.37  \\
Cross-modal coverage  & 4.29 & 4.28  \\
Naturalness \& clarity & 4.48 & 4.33  \\
\midrule
Average score & 4.41 & 4.33  \\
Preference rate & 49\% & 47\%  \\
\bottomrule
\end{tabular}
}
\caption{Summary of human evaluation comparing human-written vs.\ generated queries.}
\end{table}
\vspace{4pt}

We conduct a small-scale human study to verify the naturalness and attribute fidelity of the generated queries. We randomly sample 100 target products from all domains. For each product, Annotator~A is given the product image, its structured metadata, and the internally extracted fine-grained attributes, and is asked to write a natural-language search query that a user might realistically issue—without access to the model-generated query.

Annotator~B then evaluates, under double-blind conditions, both the human-written and model-generated queries. Each query is assessed on three dimensions using a 1--5 Likert scale: (i) \emph{attribute correctness}, (ii) \emph{cross-modal coverage}, and (iii) \emph{naturalness and clarity}. Annotator~B additionally selects which query better matches the target product.

Overall, model-generated queries achieve scores comparable to human-written ones across all criteria, with only a small gap in naturalness. The near-equal preference distribution indicates no strong annotator bias toward either source. These results confirm that the proposed generation pipeline produces high-quality, multi-condition queries suitable for cross-modal retrieval research.

\end{document}